\documentclass[sigconf]{acmart}
\definecolor{xianquan}{RGB}{255,0,255}

\usepackage{multirow}
\usepackage{algorithm}
\usepackage{algorithmic}
\usepackage{graphicx}  
\usepackage[table]{xcolor}
\usepackage{placeins} 
\AtBeginDocument{%
  }

\setcopyright{acmlicensed}
\copyrightyear{2018}
\acmYear{2018}
\acmDOI{XXXXXXX.XXXXXXX}
\acmConference[KDD'26]{SIGKDD}{August 9--August 13, 2026}{Jeju, Korea}

\acmISBN{978-1-4503-XXXX-X/2018/06}

\author{Jintao Zhang}
\email{zjttt@mail.ustc.edu.cn}
\affiliation{%
  \institution{State Key Laboratory of Cognitive Intelligence, University of Science and Technology of China}
  \city{Hefei}
  \state{Anhui}
  \country{China}
}

\author{Mingyue Cheng}
\authornote{Corresponding author.}
\email{mycheng@ustc.edu.cn}
\affiliation{%
  \institution{State Key Laboratory of Cognitive Intelligence, University of Science and Technology of China}
  \city{Hefei}
  \state{Anhui}
  \country{China}
}

\author{Zirui Liu}
\email{liuzirui@mail.ustc.edu.cn}
\affiliation{%
  \institution{State Key Laboratory of Cognitive Intelligence, University of Science and Technology of China}
  \city{Hefei}
  \state{Anhui}
  \country{China}
}

\author{Xianquan Wang}
\email{wxqcn@mail.ustc.edu.cn}
\affiliation{%
  \institution{State Key Laboratory of Cognitive Intelligence, University of Science and Technology of China}
  \city{Hefei}
  \state{Anhui}
  \country{China}
}

\author{Yitong Zhou}
\email{yitong.zhou@mail.ustc.edu.cn}
\affiliation{%
  \institution{The First Affiliated Hospital of University of Science and Technology of China}
  \city{Hefei}
  \state{Anhui}
  \country{China}
}

\author{Qi Liu}
\email{qiliuql@ustc.edu.cn}
\affiliation{%
  \institution{State Key Laboratory of Cognitive Intelligence, University of Science and Technology of China}
  \city{Hefei}
  \state{Anhui}
  \country{China}
}

\begin{document}

\title{Towards Stable and Structured Time Series Generation with Perturbation-Aware Flow Matching}


\renewcommand{\shortauthors}{Trovato et al.}

\begin{abstract}
Time series generation is critical for a wide range of applications, which greatly supports downstream analytical and decision-making tasks. However, the inherent temporal heterogeneous induced by localized perturbations present significant challenges for generating structurally consistent time series. While flow matching provides a promising paradigm by modeling temporal dynamics through trajectory-level supervision, it fails to adequately capture abrupt transitions in perturbed time series, as the use of globally shared parameters constrains the velocity field to a unified representation. To address these limitations, we introduce \textbf{PAFM}, a \textbf{P}erturbation-\textbf{A}ware \textbf{F}low \textbf{M}atching framework that models perturbed trajectories to ensure stable and structurally consistent time series generation. The framework incorporates perturbation-guided training to simulate localized disturbances and leverages a dual-path velocity field to capture trajectory deviations under perturbation, enabling refined modeling of perturbed behavior to enhance the structural coherence. In order to further improve sensitivity to trajectory perturbations while enhancing expressiveness, a mixture-of-experts decoder with flow routing dynamically allocates modeling capacity in response to different trajectory dynamics. Extensive experiments on both unconditional and conditional generation tasks demonstrate that PAFM consistently outperforms strong baselines. Code is available at \url{https://anonymous.4open.science/r/PAFM-03B2}.

\end{abstract}



\received{20 February 2007}
\received[revised]{12 March 2009}
\received[accepted]{5 June 2009}

\maketitle

\section{Introduction}

\begin{figure}[htpb]
    \centering    
    \includegraphics[width=\linewidth]{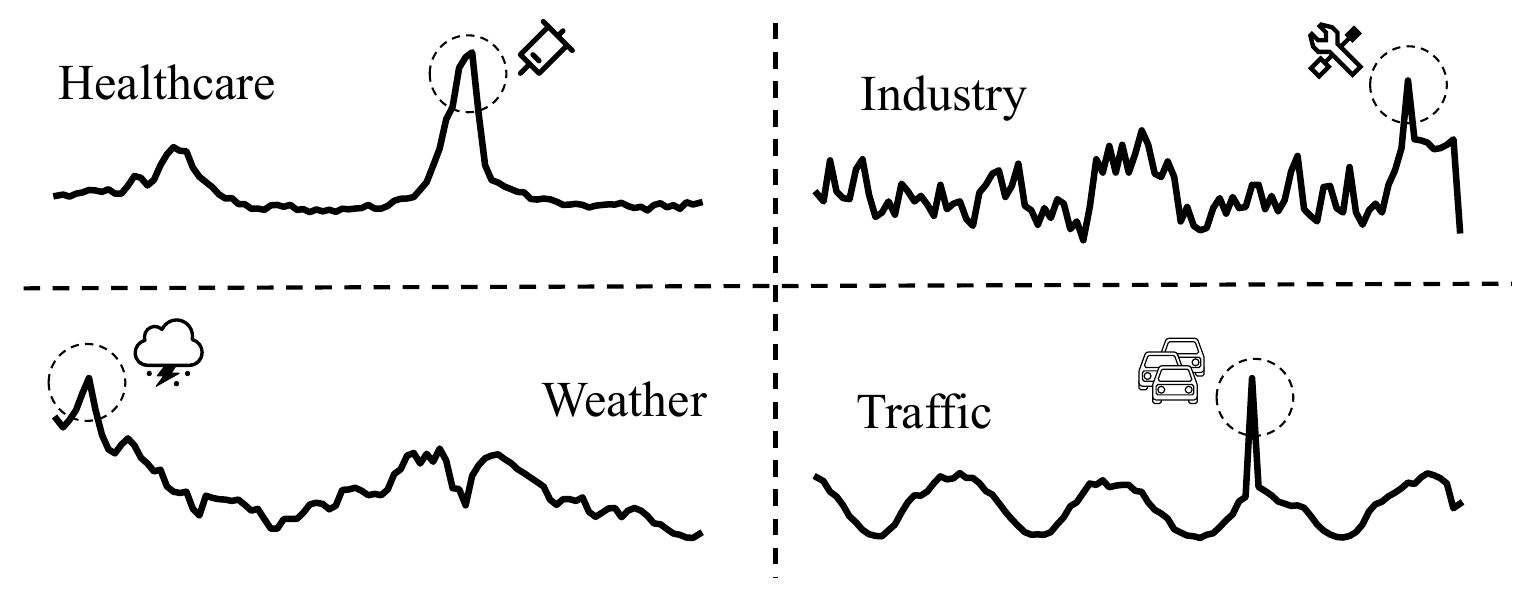}
    \vspace{-0.15in}
    \caption{Time series are inherently heterogeneous, as localized perturbations and abrupt shifts disrupt structural temporal continuity and hinder stable generation.}
    \label{fig:perturbation_examples}
    \vspace{-0.25in}
\end{figure}

\begin{figure*}[htpb]
    \centering    
    \includegraphics[width=\linewidth]{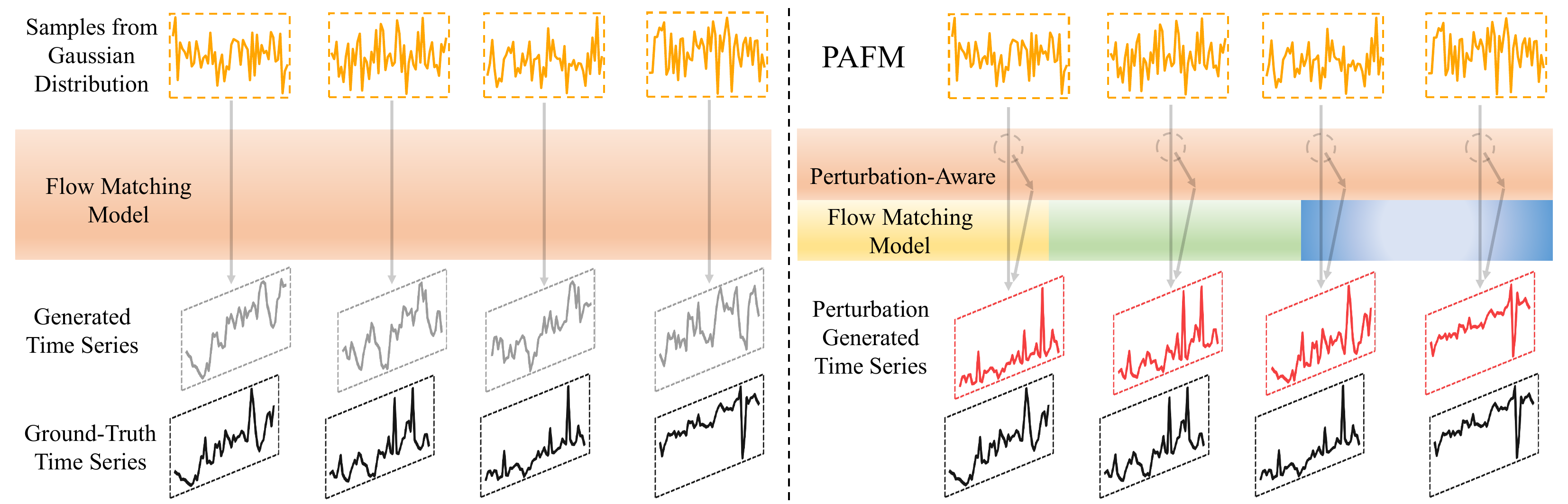}
    \vspace{-0.15in}
    \caption{Comparison between standard Flow Matching (left) and our PAFM framework (right). Standard FM fails to account for localized perturbations, resulting in misaligned or structurally inconsistent trajectories. In contrast, PAFM employs a dual-path design that integrates perturbation-guided refinement and a structure-aware decoder to adapt the velocity field, enabling temporally coherent and structurally consistent generation.}
    \label{fig:dj}
    \vspace{-0.1in}
\end{figure*}

Time series analysis plays a crucial role in domains such as finance~\cite{liuyan1}, industry~\cite{y1}, and traffic systems~\cite{y2}. However, its effectiveness is hindered by data scarcity and the rarity of critical temporal patterns~\cite{luo2025time, wang2025can}. To mitigate these challenges, researchers have explored the generation of realistic synthetic time series to facilitate downstream tasks such as forecasting~\cite{y1111} and imputation~\cite{y111}. As illustrated in Figure~\ref{fig:perturbation_examples}, real-world time series are inherently heterogeneous~\cite{cheng2024hmf}, primarily due to localized perturbations that cause abrupt structural changes and statistical shifts in temporal dynamics. For example, in intensive care units, clinical interventions often induce abrupt local fluctuations in patient physiological series~\cite{liuyan2}. These perturbations introduce substantial difficulties for generative models, which often fail to capture abrupt regime transitions and yield structurally inconsistent or unrealistic outputs~\cite{y4}.

Various generative frameworks have been explored to capture the complex dynamics of time series, including Generative Adversarial Networks~\cite{gan}, Variational Autoencoders~\cite{vAE}, and diffusion-based probabilistic models~\cite{adalayer1}. While effective at modeling marginal distributions~\cite{diffusion-ts, timevae, GAN6}, these methods typically rely on globally aggregated latent representations and implicit generation mechanisms, which limit their ability to capture localized structural variations and perturbation behaviors. Flow Matching (FM)~\cite{flow} has recently emerged as a principled generative paradigm by introducing explicit supervision over continuous trajectories via velocity field modeling. FM learns a velocity field that continuously transports samples drawn from a Gaussian prior toward target time series along straight line trajectories, enabling efficient training and fast inference during the generative process.


However, existing FM methods face critical limitations when applied to perturbation time series. As illustrated in the left panel of Figure~\ref{fig:dj}, the generated trajectories often fail to align with dynamic perturbations in the target series, revealing their inability to handle inherent heterogeneity effectively. This inadequacy stems from three fundamental issues. First, most FM models are trained without perturbation-aware supervision, leaving the velocity field insensitive to dynamic changes and localized deviations. Second, most FM models rely on globally shared velocity parameters~\cite{confm, y6}, which enforce temporal homogeneity and limit adaptability to perturbed patterns. Third, their feedforward networks~\cite{csdi, diffusion-ts, y7} lack the inductive bias required to distinguish stable from perturbed trajectories, resulting in insufficient modeling of localized transitions.

To address the aforementioned challenges, we propose \textbf{PAFM}, a \textbf{P}erturbation-\textbf{A}ware Structure-Sensitive \textbf{F}low \textbf{M}atching framework. As illustrated in the right panel of Figure~\ref{fig:dj}, PAFM integrates a perturbation-guided training strategy with a structure-aware decoder to improve adaptability under dynamic regime transitions. Specifically, we introduce a Trajectory Simulation Module that injects localized perturbations to simulate potential disruptions. These perturbations enhance sensitivity to local dynamics and improve velocity alignment. To enhance perturbation awareness and ensure structural consistency, we design a Dual-Path Velocity Field that jointly models original and perturbed trajectory series. In parallel, we construct a Trajectory Decoder equipped with a Flow Routing Mixture-of-Experts. This module dynamically routes trajectory segments based on velocity patterns, assigning different temporal regions to specialized experts. Finally, the Velocity Refinement Module adjusts the estimated velocity and introduces targeted supervision on flow trajectories to improve velocity alignment.

Our contributions are summarized as follows:

\begin{itemize}
    \item We propose \textbf{PAFM}, a perturbation-aware flow matching framework that injects localized trajectory perturbations to simulate dynamic transitions and jointly models original and perturbed trajectories via a Dual-Path Velocity Field, enabling the model to leverage perturbation responses for velocity refinement.
    \item We design a trajectory-level mixture-of-experts decoder that dynamically routes segments based on local velocity patterns, allowing the model to focus capacity on structurally unstable regions.
    \item Extensive experiments on both unconditional and conditional time series generation tasks demonstrate that PAFM consistently outperforms state-of-the-art baselines across multiple real-world datasets.
\end{itemize}

\section{Related Work}

\subsection{Time series generation}

Time series generation is inherently challenging due to complex temporal dependencies, irregular sampling, and non-stationary dynamics. Early efforts predominantly relied on generative adversarial networks (GANs), such as TimeGAN~\cite{timegan}, which integrates supervised losses and temporal embeddings to preserve sequential structure. Later advances like PSA-GAN~\cite{GAN5} and ITF-GAN~\cite{GAN6} introduce attention-based modeling and functional decomposition to improve long-range generation and interpretability. Beyond GANs, TimeVAE~\cite{timevae} leverages hierarchical latent variables for structured representation, while Hypertime~\cite{hypertime} synthesizes series through hypernetwork-based interpolation.

More recently, diffusion probabilistic models have gained traction as a compelling alternative~\cite{cheng2025comprehensive, zhang2024conditional}, demonstrating superior sample quality and enhanced training stability compared to GAN-based methods. TimeGrad~\cite{DIFF2}, CSDI~\cite{csdi}, and SSSD~\cite{DIFF4} extend diffusion-based frameworks to conditional generation and missing data imputation, particularly in settings with irregularly sampled inputs, leveraging self-supervised masking techniques. For unconditional generation, TimeDiff~\cite{DIFF5} employs the non-autoregressive approach to reduce boundary artifacts, while recent works such as TSDE~\cite{TSDE} and Diffusion-TS~\cite{diffusion-ts} incorporate structured inductive biases to better capture temporal patterns. Despite recent progress, these models depend on stochastic sampling or latent variables, lacking mechanisms for direct and interpretable trajectory control.


\subsection{Flow Matching}

Flow Matching (FM) has recently emerged as a prominent approach in generative modeling. By explicitly learning the velocity field of continuous-time transformations, FM enables the construction of deterministic generation trajectories, leading to improved inference efficiency and controllability. FM~\cite{flow} and RF~\cite{flow2} pioneered this framework by formulating conditional probability paths based on optimal transport theory and directly fitting generation trajectories through velocity field learning. 


FM has demonstrated strong performance across a wide range of domains, particularly in multimodal generation tasks. In image synthesis, RFT~\cite{flow3} introduces geometry-aware flow paths that condition the velocity field on object layout and segmentation priors, enabling regionally coherent synthesis and fine-grained structural alignment. In audio generation, LAFMA~\cite{flow4} builds high-fidelity generation pipelines from waveforms. In symbolic and molecular domains such as protein design and DNA sequence generation, Dirichlet Flow Matching~\cite{flow5} address the challenge of modeling discrete, non-differentiable trajectories of DNA. In real-world applications including human motion synthesis, financial decision-making, and recommendation systems, methods such as Motion Flow Matching~\cite{flow6}, FlowHFT~\cite{flow8}, GFN4Rec~\cite{gfn4rec} and GFN4Retention~\cite{GFN4Retention} achieve stable and efficient trajectory generation through conditional modeling and structure-aware designs.

FM has also shown promising potential in time series modeling. TFM~\cite{flow9} introduces trajectory supervision to address irregularity and uncertainty in clinical series. CFMTS~\cite{flow10} integrates FM with Neural ODEs to model continuous conditional probability paths. TSflow~\cite{y6} further incorporates Gaussian process priors to enhance structural guidance and predictive robustness. These advances highlight the broad applicability of FM in time series tasks. However, most existing FM-based models primarily focus on trajectory supervision or prior guidance, while lacking explicit mechanisms to account for local perturbation responses and structural heterogeneity, which are critical in complex temporal dynamics.

\begin{figure*}[htpb]
    \centering    
    \includegraphics[width=\linewidth]{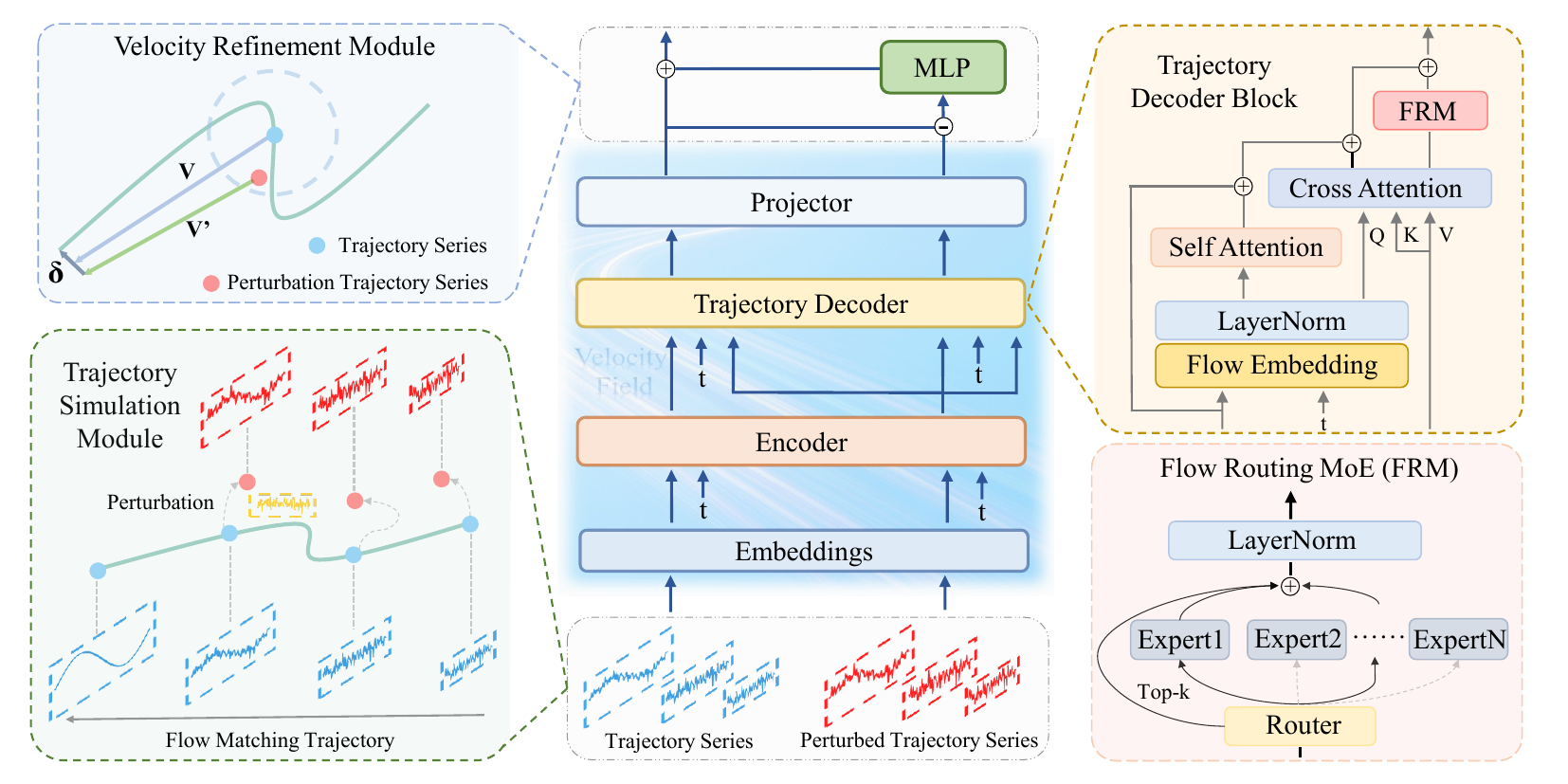}
    \vspace{-0.1in}
    \caption{Overview of the proposed perturbation-aware flow framework. (1) Trajectory Perturbation Block, consisting of Trajectory Simulation Module and Velocity Refinement Module, introduces localized perturbations to simulate abrupt disturbances and leverages dynamic feedback to refine the velocity field;  (2) Velocity Field with dual-path models perturbation-aware dynamics;  (3) Trajectory Decoder Block incorporates the Flow Routing Mixture-of-Experts (FRM) for structural representation.}
    \label{fig:mainpic}
    \vspace{-0.1in}
\end{figure*}

\vspace{-0.05in}

\section{Methodology}

\subsection{Problem Formulation}

A multivariate time series is defined as \( x_1 \in \mathbb{R}^{\tau \times d} \), where \( \tau \) denotes the time steps and \( d \) is the dimensionality of observations at each step. Given a dataset \( \mathcal{D}_A = \{x_1^{(i)}\}_{i=1}^N \) consisting of \( N \) time series samples, our goal is to build a framework based on Flow Matching (FM) that maps Gaussian noise to high-quality synthetic series.

Specifically, FM aims to learn a vector field \( v_\theta(x, t) \) and define a continuous trajectory through the ordinary differential equation:
\begin{equation}
\frac{d\gamma_{x_1}(t)}{dt} = v_\theta(\gamma_{x_1}(t), t), \quad \gamma_{x_1}(0) = x_0,
\end{equation}
where \( \gamma_{x_1}(t) \) denotes the trajectory starting from an initial noise sample \( x_0 \sim \mathcal{N}(0, \boldsymbol{I}) \) and evolving toward the series \( x_1 \sim P_\text{data} \). At each time step \( t \in [0, 1] \), we define \( x_t = \gamma_{x_1}(t) \) as the trajectory series of FM. To align the learned flow with the underlying dynamics of the target distribution, FM defines a ground-truth vector field \( u(x, t) \) and its associated probability path \( P_t \), and formulates the training objective as the minimization of the following loss:
\begin{equation}
\mathbb{E}_{t, P_t} \left\| v_\theta(x_t, t) - u(x_t, t) \right\|_2^2.
\end{equation}

However, directly solving this objective is computationally intractable, as both \( u \) and \( p_t \) are implicitly defined through the continuity equation. Conditional Flow Matching~\cite{flow} addresses this by conditioning on a target sample \( x_1 \sim P_\text{data} \) and regressing \( v_\theta(x_t, t) \) against the conditional vector field \( u(x_t, t\mid x_1) \):

\begin{equation}
\mathbb{E}_{t, P_1(x_1)} \mathbb{E}_{P_t(x_t \mid x_1)} \left\| v_\theta(x_t, t) - u(x_t, t \mid x_1) \right\|_2^2.
\end{equation}

\subsection{Structure Overview}

Our framework integrates perturbation modeling with expert-driven decoding to enable stable and structure coherent time series generation. The framework consists of four components: (1) \textbf{Trajectory Perturbation Block} comprises a Trajectory Simulation Module, which introduces localized perturbations into the trajectory to simulate abrupt disturbances, and a Velocity Refinement Module, which leverages dynamic feedback to guide velocity field refinement. (2) \textbf{Velocity Field}, utilizing a dual-path architecture to simultaneously capture the dynamics of both original and perturbed trajectories. (3) \textbf{Trajectory Decoder}, composed of multiple blocks and integrating a Flow Routing Mixture-of-Experts (FRM) module to support structurally-aware modeling of diverse temporal dynamics. (4) \textbf{Trajectory Optimization Block}, which applies direct supervision to refine velocity alignment with the trajectory.

\subsection{Trajectory Perturbation Block}

To improve the ability of the model to capture directional flow within local neighborhoods, we propose the Trajectory Perturbation Block, which comprises two components: the \textit{Trajectory Simulation Module} and the \textit{Velocity Refinement Module}. These modules jointly construct perturbation-aware representations that simulate perturbed series and locally refine the velocity field.

In the \textit{Trajectory Simulation Module}, we inject a small Gaussian perturbation \( \epsilon \sim \mathcal{N}(0, I_d) \) into the trajectory series \( x_t \in \mathbb{R}^{\tau \times d} \) to obtain a perturbed sample \( x_t' = x_t + \sigma \cdot \epsilon \), where \( \sigma \) controls the perturbation scale. This injection is intended to probe how the velocity field responds to local input variations, especially in regions near dynamical transitions. By evaluating the predicted velocity at both the original and perturbed inputs, we define the perturbation-induced response as:
\begin{equation}
\delta(x_t, t) = v_\theta(x_t', t) - v_\theta(x_t, t).
\end{equation}

Assuming that the velocity field \( v_\theta(x, t) \) is differentiable with respect to \( x \), we can apply the first-order Taylor expansion around \( x_t \) under small \( \sigma \), yielding:
\begin{equation}
v_\theta(x_t + \sigma \cdot \epsilon, t) \approx v_\theta(x_t, t) + \sigma \cdot \nabla_{x_t} v_\theta(x_t, t) \cdot \epsilon.
\end{equation}

Substituting this into the definition of \( \delta(x_t, t) \), we obtain:
\begin{equation}
\delta(x_t, t) \approx \sigma \cdot \nabla_{x_t} v_\theta(x_t, t) \cdot \epsilon.
\end{equation}

This expression reveals that \( \delta(x_t, t) \) approximates a Jacobian-vector product, which represents a directional derivative of the velocity field along the perturbation direction. Therefore, \( \delta(x_t, t) \) captures how sensitive the velocity prediction is to local directional variations—effectively encoding first-order geometric information of the flow manifold around \( x_t \).

In the \textit{Velocity Refinement Module}, we leverage this sensitivity measure to refine the predicted velocity field. Specifically, we treat \( \delta(x_t, t) \) as a directionally-aware structural correction term and incorporate it into the final velocity prediction:
\begin{equation}
v_{\text{final}}(x_t, t) = v_\theta(x_t, t) + \alpha \cdot \delta(x_t, t) \cdot w(v_\theta(x_t, t)),
\end{equation}
where \( \alpha \) is a scalar hyperparameter, and \( w(\cdot) \) is a learnable gating function, which modulates the correction strength based on the current velocity output. This correction mechanism adaptively linearizes the velocity field around the input \( x_t \), guided by local perturbation responses. Incorporating Jacobian-based adjustments enables the model to capture localized nonlinear dynamics and enhance trajectory alignment in unstable regions.

\subsection{Velocity Field}
The velocity field \( v_\theta \) incorporates explicit perturbation-awareness by modeling both the original and perturbed trajectories through a dual-path design. This enables dynamically stable and  structurally consistent generation, supporting high-quality time series synthesis under varying conditions.

To realize this, \( v_\theta \) is implemented using a convolutional embedding layer, a transformer encoder, a trajectory-attentive decoder, and an output projector. It jointly encodes both the original and perturbed trajectories \( x_t \) and \( x_t' \), enabling the velocity field to capture the impact of local perturbations on global directional dynamics.

For \( x_t \in \mathbb{R}^{\tau \times d} \), it is transformed into a hidden representation \( H_t \in \mathbb{R}^{\tau \times d_{\text{model}}} \) via a convolutional embedding layer:
\begin{equation}
H_t = \text{Embedding}(x_t).
\end{equation}

The representation \( H_t \) is then encoded by a transformer stack to produce the contextual representation \( E_t \):
\begin{equation}
E_t = \text{Encoder}(H_t).
\end{equation}

An identical pipeline is applied to the perturbed series \( x_t' \), resulting in its corresponding hidden state \( h_t' \) and contextual representation \( E_t' \). To align perturbed trajectories with the intrinsic dynamics of their clean counterparts, the model adopts a trajectory attentive decoding strategy that simultaneously exploits contextual representations from both trajectory and perturbed series.

The decoder generates the clean trajectory output as:
\begin{equation}
D_t = \text{Decoder}(E_t, t, E_t'),
\end{equation}
where \( E_t \) encodes the clean trajectory context, and \( E_t' \) offers auxiliary guidance from its perturbed counterpart. In parallel, the perturbed trajectory is decoded through:
\begin{equation}
D_t' = \text{Decoder}(E_t', t, E_t').
\end{equation}

Finally, the decoder output \( D_t \) is projected back to the original input dimension through a convolutional layer:
\begin{equation}
v_\theta(x_t) = \text{Projector}(D_t).
\end{equation}

Meanwhile, \( D_t' \) is processed analogously to obtain \( v_\theta(x_t') \).

\subsection{Trajectory Decoder}
We introduce a Trajectory Decoder composed of multiple blocks, equipped with FRM module that dynamically adjusts modeling capacity in response to diverse trajectory dynamics. FRM adaptively routes trajectory segments to expert networks through a gating mechanism informed by the local temporal features, allowing adaptive modeling in structurally complex regions. This structure specialization~\cite{moe, deepseekmoe} enhances the expressiveness of the velocity field \( v_\theta \) and improves both structural temporal consistency and sensitivity to abrupt shifts.

Given the encoder representations \( E^1 , E^2 \in \mathbb{R}^{d_\text{model} \times d} \), and the current timestep \( t \), the decoder first applies a flow embedding module with Adaptive Layer Normalization (AdaLN)~\cite{adalayer1, adalayer2} to obtain temporally-aware features. The representations are then processed by self-attention and cross-attention layers to extract temporal and contextual dependencies, followed by the FRM module, which adaptively aggregates expert outputs based on local trajectory patterns.

To better capture dynamic patterns across time while preserving temporal structure, the FRM module employs a trajectory-level gating mechanism that adaptively routes contextual representations to specialized experts based on localized temporal dynamics. Within the FRM module, the input representation obtained after the attention mechanism is denoted as \( Z \in \mathbb{R}^{\tau \times d_{\text{model}}} \). A routing score matrix \( G \in \mathbb{R}^{\tau \times M} \) is then computed, where \( G_{i, m} \) indicates the relevance between the \( i \)-th time step and the \( m \)-th expert. The gating scores \( G \) are obtained through the trainable linear projection:

\begin{equation}
G = Z\textbf{W}^\top,
\end{equation}
where \( \textbf{W} \in \mathbb{R}^{M \times d_{\text{model}}} \) is the trainable gating weight matrix.  

At each position along the series, we select the top-$k$ experts with the highest gating scores using a sorting operator \( \pi \). Let \( \mathcal{I} = \pi(G) \in \mathbb{R}^{\tau \times k} \) denote the indices of the top-$k$ experts for each time step, and let the corresponding gating scores be \( G_{\mathcal{I}} \in \mathbb{R}^{\tau \times k} \). These top-$k$ scores are normalized using softmax:
\begin{equation}
g^{(m)} = \mathrm{Softmax}(G_{\mathcal{I}})^{(m)}, \quad m = 1, \dots, k,
\end{equation}
where \( g^{(m)} \in \mathbb{R}^{\tau \times 1} \) denotes the normalized gating weight for expert \( m \) across all positions.

Each selected expert \( m \) applies a feedforward transformation \( f_m(\cdot) \) to the representation \( Z \in \mathbb{R}^{\tau \times d_{\text{model}}} \), and the result is modulated by its corresponding routing weight:
\begin{equation}
y^{(m)} = g^{(m)} \odot f_m(Z),
\end{equation}
where \( \odot \) denotes element-wise multiplication with broadcasting along feature dimension. The aggregated output is computed as:
\begin{equation}
y = \sum_{m=1}^{k} y^{(m)}.
\end{equation}

A residual connection followed by layer normalization is applied to stabilize the output:
\begin{equation}
Z^{\text{out}} = \mathrm{LayerNorm}(Z + y),
\end{equation}
where \( y \in \mathbb{R}^{\tau \times 1} \) is broadcast along the hidden dimension to match the shape of \( Z \in \mathbb{R}^{\tau \times d_{\text{model}}} \).

This expert-aware decoding mechanism enables the FRM module to dynamically specialize across localized flow patterns while maintaining global temporal coherence.

\subsection{Trajectory Optimization Block}

Building upon the refined velocity output \( v_{\text{final}}(x_t, t) \) obtained from the perturbation-aware correction process, we adopt it as the supervision target and define a trajectory-level objective to optimize its alignment with the underlying temporal dynamics. To encourage directional consistency and structural fidelity between the perturbed and original series, we introduce the following loss:


\begin{equation}
\mathcal{L}_{\text{total}} = \left\| v_{\text{final}}(x_t, t) - (x_1 - x_0) \right\|_2^2,
\end{equation}

By directly supervising \( v_{\text{final}} \), the model can better exploit structural cues embedded in the perturbation series and achieve improved trajectory-velocity alignment.

\section{Experiments}
To comprehensively evaluate the effectiveness of our proposed model, we employ the six benchmark datasets and conduct a series of experiments under both unconditional and conditional generation settings to assess the quality of the generated series. Our method is compared against seven state-of-the-art baselines.

\begin{table}[thb]\centering
    \vspace{-0.1in}
    \caption{Overview of the dataset characteristics.}
    \vspace{-0.1in}
    \label{tab1}%
    \resizebox{0.48\textwidth}{!}{
    \large
    \begin{tabular}{c|cccccc}
    \toprule
    Dataset & Sines & Stocks & ETTh1 & MujoCo & Energy & fMRI \\
    \midrule
    \#Samples & 10000 & 3773  & 17420 & 10000 & 19711 & 10000 \\
    Dim   & 5     & 6     & 7     & 14    & 28    & 50 \\
    \bottomrule
    \end{tabular}
    }
    \vspace{-0.2in}
\end{table}

\paragraph{Dataset.}
We evaluate our proposed method on six datasets, including four real-world and two synthetic benchmarks, following Diffusion-TS~\cite{diffusion-ts}, as summarized in Table~\ref{tab1}. The \textbf{Stocks}\footnote{\url{https://finance.yahoo.com/quote/GOOG}} dataset contains daily Google stock prices from 2004 to 2019, with six financial features. The \textbf{ETTh1}\footnote{\url{https://github.com/zhouhaoyi/ETDataset}}~\cite{informer} dataset records power transformer measurements, including electrical load and oil temperature, at 15-minute intervals. The \textbf{Energy}\footnote{\url{https://archive.ics.uci.edu/dataset/374/energy}} dataset from the UCI repository provides 28-dimensional measurements related to appliance energy consumption. The \textbf{fMRI}\footnote{\url{https://www.fmrib.ox.ac.uk/datasets/netsim/}} dataset consists of simulated BOLD time series; we use a variant with 50 variables. Among the synthetic datasets, \textbf{Sines}\footnote{\url{https://github.com/jsyoon0823/TimeGAN}}~\cite{timegan} comprises five series generated with varying frequencies and phases, and \textbf{MuJoCo}\footnote{\url{https://github.com/google-deepmind/dm_control}} is a 14-dimensional time series derived from physics-based simulations.

\begin{table*}[htbp]
  \centering
  \caption{Unconditional Time Series Generation performance across six datasets. Bold indicates the best performance, and underlining indicates the second-ranking result.}
    \vspace{-0.05in}
    \begin{tabular}{c|ccccccc}
    \toprule
    Metric & Methods & Sines & Stocks & ETTh1  & MuJoCo & Energy & fMRI \\
    \midrule
    \multirow{7}[0]{*}{\begin{tabular}[c]{@{}c@{}}Discriminative\\Score\\ (Lower Better)\end{tabular}} & \cellcolor{lightgray!30}\textbf{Ours} & \cellcolor{lightgray!30}\textbf{0.006±0.001} & \cellcolor{lightgray!30}\textbf{0.021±0.017} & \cellcolor{lightgray!30}\textbf{0.008±0.005} & \cellcolor{lightgray!30}\textbf{0.007±0.004} & \cellcolor{lightgray!30}\textbf{0.047±0.006} & \cellcolor{lightgray!30}\underline{0.243±0.048} \\
          & Diffusion-TS & \underline{0.006±0.007}  & \underline{0.067±0.015}  & \underline{0.061±0.009} & \underline{0.008±0.002} & \underline{0.122±0.003} & \textbf{0.167±0.023} \\
          & TimeGAN & 0.011±0.008 & 0.102±0.021 & 0.114±0.055 & 0.238±0.068 & 0.236±0.012 & 0.484±0.042 \\
          & TimeVAE & 0.041±0.044 & 0.145±0.120  & 0.209±0.058 & 0.230±0.102 & 0.499±0.000  & 0.476±0.044 \\
          & DiffWave & 0.017±0.008 & 0.232±0.061 & 0.190±0.008 & 0.203±0.096 & 0.493±0.004 & 0.402±0.029 \\
          & DiffTime & 0.013±0.006 & 0.097±0.016 & 0.100±0.007 & 0.154±0.045 & 0.445±0.004 & 0.245±0.051 \\
          & Cot-GAN & 0.254±0.137 & 0.230±0.016  & 0.325±0.099 & 0.426±0.022 & 0.498±0.002 & 0.492±0.018 \\
    \midrule
    \multirow{7}[0]{*}{\begin{tabular}[c]{@{}c@{}}Predictive\\Score\\ (Lower Better)\end{tabular}} & \cellcolor{lightgray!30}\textbf{Ours} & \cellcolor{lightgray!30}\textbf{0.093±0.000} & \cellcolor{lightgray!30}\textbf{0.036±0.000 } & \cellcolor{lightgray!30}\underline{0.120±0.006} & \cellcolor{lightgray!30}\underline{0.009±0.001} & \cellcolor{lightgray!30}\textbf{0.250±0.000 } & \cellcolor{lightgray!30}\textbf{0.099±0.000} \\
          & Diffusion-TS & \textbf{0.093±0.000} & \textbf{0.036±0.000 } & \textbf{0.119±0.002} & \textbf{0.007±0.000} & \textbf{0.250±0.000 } & \textbf{0.099±0.000} \\
          & TimeGAN & 0.093±0.019 & 0.038±0.001  & 0.124±0.001 & 0.025±0.003 & 0.273±0.004 & 0.126±0.002 \\
          & TimeVAE & \textbf{0.093±0.000} & 0.039±0.000 & 0.126±0.004 & 0.012±0.002 & 0.292±0.000 & 0.113±0.003 \\
          & DiffWave & \textbf{0.093±0.000} & 0.047±0.000 & 0.130±0.001 & 0.013±0.000 & 0.251±0.000 & 0.101±0.000 \\
          & DiffTime & \textbf{0.093±0.000} & 0.038±0.001 & 0.121±0.004 & 0.010±0.001 & 0.252±0.000 & 0.100±0.000 \\
          & Cot-GAN & 0.100±0.000 & 0.047±0.001 & 0.129±0.000 & 0.068±0.009 & 0.259±0.000 & 0.185±0.003 \\
    \midrule
    \multirow{7}[0]{*}{\begin{tabular}[c]{@{}c@{}}Context-FID\\Score\\ (Lower Better)\end{tabular}} & \cellcolor{lightgray!30}\textbf{Ours} & \cellcolor{lightgray!30}\textbf{0.005±0.001} & \cellcolor{lightgray!30}\textbf{0.025±0.001} & \cellcolor{lightgray!30}\textbf{0.022±0.002} & \cellcolor{lightgray!30}\underline{0.017±0.003} & \cellcolor{lightgray!30}\textbf{0.027±0.002} & \cellcolor{lightgray!30}0.276±0.011 \\
          & Diffusion-TS & \underline{0.006±0.000} & 0.147±0.025 & \underline{0.116±0.010} & \textbf{0.013±0.001} & \underline{0.089±0.024} & \textbf{0.105±0.006} \\
          & TimeGAN & 0.101±0.014 & \underline{0.103±0.013} & 0.300±0.013 & 0.563±0.052 & 0.767±0.103 & 1.292±0.218 \\
          & TimeVAE & 0.307±0.060 & 0.215±0.035 & 0.805±0.186 & 0.251±0.015 & 1.631±0.142 & 14.449±0.969 \\
          & DiffWave & 0.014±0.002 & 0.232±0.032 & 0.873±0.061 & 0.393±0.041 & 1.031±0.131 & \underline{0.244±0.018} \\
          & DiffTime & 0.006±0.001 & 0.236±0.074 & 0.299±0.044 & 0.188±0.028 & 0.279±0.045 & 0.340±0.015 \\
          & Cot-GAN & 1.337±0.068 & 0.408±0.086 & 0.980±0.071 & 1.094±0.079  & 1.039±0.028  & 7.813±0.550 \\
    \midrule
    \multirow{7}[0]{*}{\begin{tabular}[c]{@{}c@{}}Correlational\\Score\\ (Lower Better)\end{tabular}} & \cellcolor{lightgray!30}\textbf{Ours} & \cellcolor{lightgray!30}\textbf{0.014±0.014} & \cellcolor{lightgray!30}\textbf{0.003±0.006} & \cellcolor{lightgray!30}\textbf{0.023±0.009} & \cellcolor{lightgray!30}\textbf{0.186±0.019} & \cellcolor{lightgray!30}\textbf{0.668±0.142} & \cellcolor{lightgray!30}\textbf{1.382±0.062} \\
          & Diffusion-TS & \underline{0.015±0.004} & \underline{0.004±0.001} & \underline{0.049±0.008}  & \underline{0.193±0.027} & \underline{0.856±0.147} & \underline{1.411±0.042} \\
          & TimeGAN & 0.045±0.010 & 0.063±0.005 & 0.210±0.006 & 0.886±0.039 & 4.010±0.104 & 23.502±0.039 \\
          & TimeVAE & 0.131±0.010 & 0.095±0.008 & 0.111±0.020 & 0.388±0.041 & 1.688±0.226 & 17.296±0.526 \\
          & DiffWave & 0.022±0.005 & 0.030±0.020 & 0.175±0.006 & 0.579±0.018 & 5.001±0.154 & 3.927±0.049 \\
          & DiffTime & 0.017±0.004 & 0.006±0.002 & 0.067±0.005  & 0.218±0.031 & 1.158±0.095 & 1.501±0.048 \\
          & Cot-GAN & 0.049±0.010 & 0.087±0.004 & 0.249±0.009 & 1.042±0.007 & 3.164±0.061  & 26.824±0.449 \\
    \bottomrule
    \end{tabular}%
  \label{tab2}%
    \vspace{-0.05in}
\end{table*}%

\paragraph{Baselines.}

We evaluate our model on both \textbf{unconditional} and \textbf{conditional} time series generation tasks. For unconditional generation, we compare against six representative baselines: Diffusion-TS~\cite{diffusion-ts}, TimeVAE~\cite{timevae}, DiffWave~\cite{diffwave}, TimeGAN~\cite{timegan}, CotGAN~\cite{cotgan}, and DiffTime~\cite{difftime}. For conditional generation, we benchmark against Diffusion-TS~\cite{diffusion-ts} and CSDI~\cite{csdi}. To assess the effect of perturbation, we conduct controlled experiments using FlowTS with and without perturbation. Details are provided in Appendix~\ref{baselines}.

\paragraph{Metrics.}

For quantitative evaluation, we adopt four commonly used metrics to assess the quality of the generated time series: \textbf{Discriminative Score}~\cite{timegan} evaluates distributional similarity by training a classifier to distinguish between real and synthetic samples. \textbf{Predictive Score}~\cite{timegan} measures the utility of synthetic data by training a model on synthetic data and testing on real data. \textbf{Context-FID Score}~\cite{GAN5} quantifies the distance between real and synthetic representations conditioned on local temporal context. \textbf{Correlational Score}~\cite{metrics4} computes the absolute error between the cross-correlation matrices of real and synthetic data to evaluate temporal dependencies. Further details are provided in Appendix~\ref{metics}.

\paragraph{Implementation.}
For \textbf{unconditional generation}, we adopt fixed generation steps across all datasets for consistent comparison. For conditional generation, we evaluate both prediction and imputation tasks, using identical prediction steps and missing ratios for all baselines. All models are trained using the Adam optimizer~\cite{adam} with the cosine noise schedule. The learning rate linearly decays from an initial value of 0.0008, with a warm-up phase over the first 500 iterations. For conditional generation, the inference process uses 200 denoising steps with $\gamma = 0.05$.  Experiments are conducted on an NVIDIA GeForce RTX 4090D GPU. Further details regarding the implementation are provided in Appendix~\ref{implementation}.

\vspace{-0.05in}
\subsection{Analysis of Trajectory Perturbation}

\begin{figure}[htpb]
    \centering    
    \includegraphics[width=\linewidth]{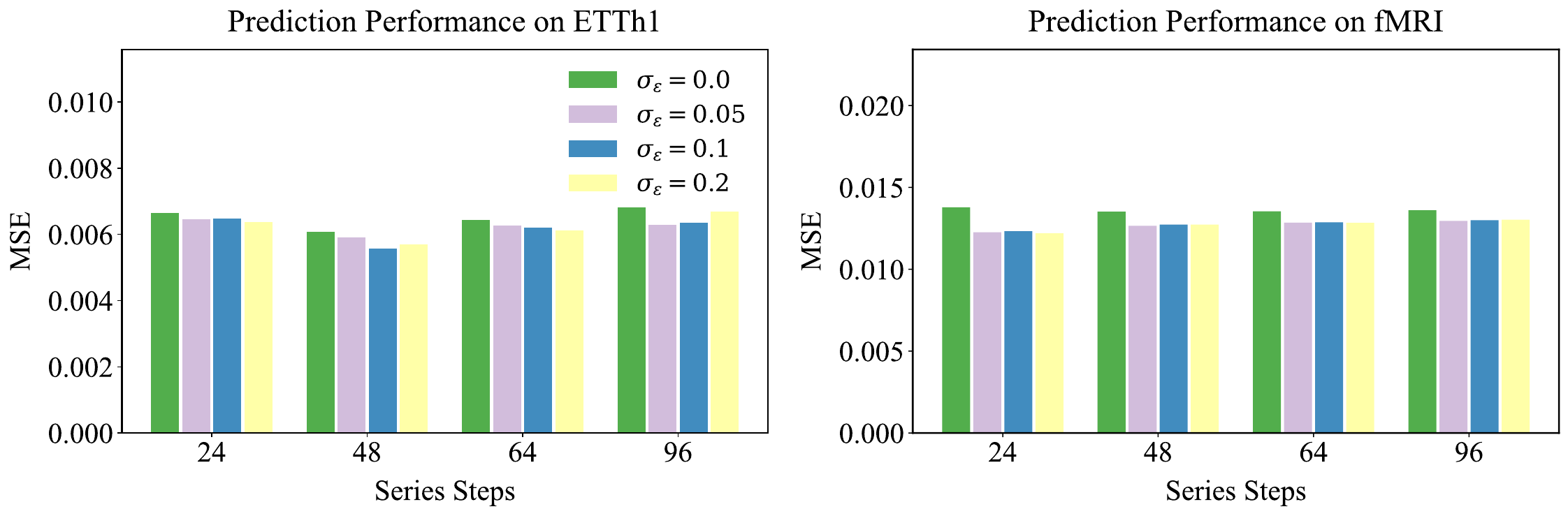}
    \vspace{-0.25in}
    \caption{Prediction performance under different perturbation magnitudes.}
    \vspace{-0.15in}
    \label{fig:f_1}
\end{figure}

\begin{figure}[htpb]
    \centering    
    \includegraphics[width=\linewidth]{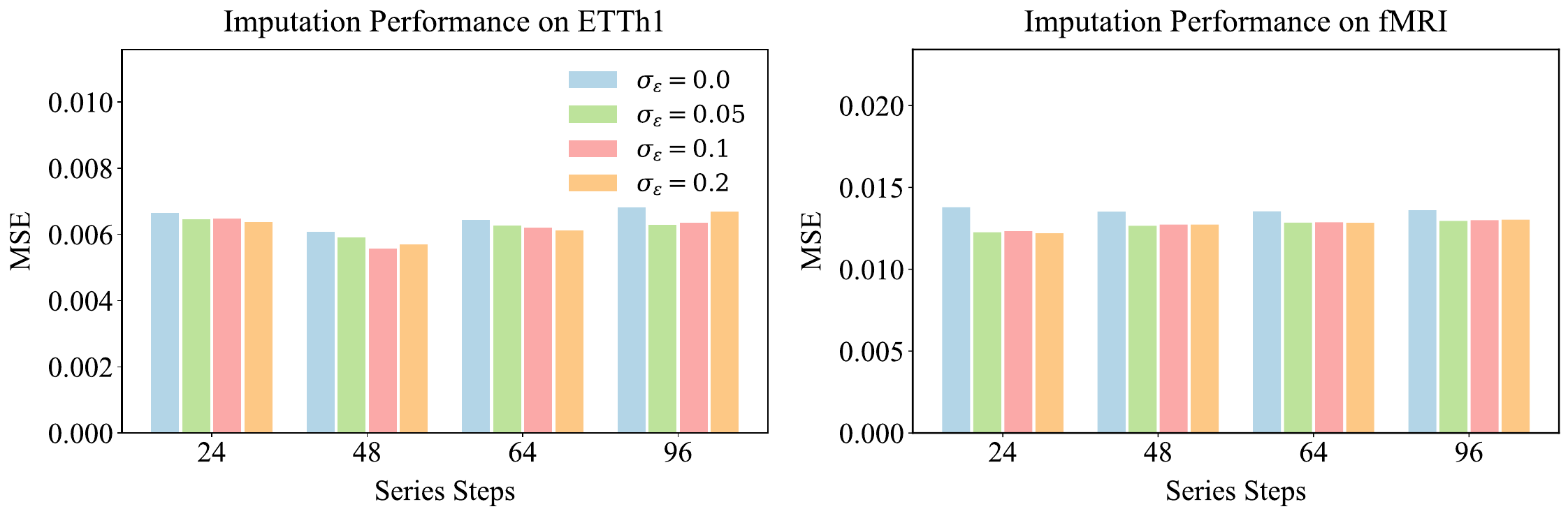}
    \vspace{-0.25in}
    \caption{Imputation performance under different perturbation magnitudes.}
    \vspace{-0.2in}
    \label{fig:i_1}
\end{figure}

To assess the impact of perturbation-based modeling, we conduct experiments on the ETTh1 and fMRI datasets under varying perturbation magnitudes and series lengths, covering both prediction and imputation tasks. As perturbation intensity increases, PAFM shows greater performance gains, especially on longer series where regime transitions are more pronounced. These results indicate that the model effectively exploits perturbation series to capture dynamic shifts associated with structural changes in perturbation time series. Results in Fig.~\ref{fig:f_1} and Fig.~\ref{fig:i_1} show that introducing perturbations enhances the sensitivity to temporal structure and improves generation consistency, particularly on longer series. By responding to small input perturbations, the model better captures local variations in the velocity field, leading to a more accurate representation of underlying dynamics. Compared to non-perturbed variants, the perturbed model achieves consistently better performance across tasks and series lengths, demonstrating that perturbation strategies offer an effective mechanism.

\paragraph{Analysis of Perturbation Method.}

\begin{figure}[htpb]
    \centering    
    \vspace{-0.05in}
    \includegraphics[width=\linewidth]{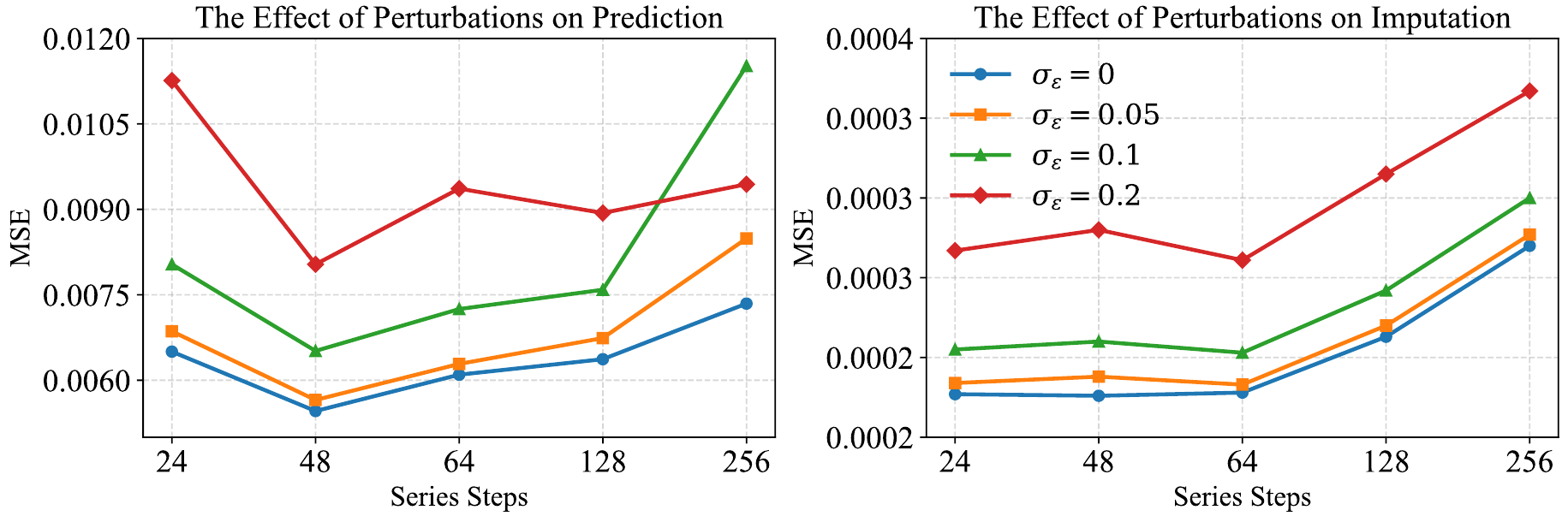}
    \vspace{-0.3in}
    \caption{Imputation and prediction under varying perturbations without perturbation training.}
    \vspace{-0.25in}
    \label{fig:a}
\end{figure}

To evaluate the effectiveness of the proposed Trajectory Perturbation Block (TPB) in conditional generation, we conduct controlled experiments using FlowTS~\cite{flowts} without perturbation-based training. Specifically, we assess 50\% time step imputation and latter-half prediction on the ETTh1 dataset under varying observed steps. As shown in Fig.~\ref{fig:a}, applying perturbations at inference without prior training leads to unstable outputs and performance degradation, indicating that perturbation of trajectory must be explicitly modeled to be effective. In contrast, Fig.~\ref{fig:f_1} and Fig.~\ref{fig:i_1} show that integrating TPB consistently improves performance, enhancing the model’s sensitivity to local directional variation and stability across conditions. These results confirm perturbation modeling as a critical component for flow-based conditional generation.

\subsection{Unconditional Time Series Generation Results}

In this section, we evaluate all methods on the 24 steps time series generation tasks, using the same benchmark settings to ensure fair and consistent comparison across models.

As shown in Table~\ref{tab2}, the proposed method demonstrates consistently strong performance in both generation quality and structural modeling for time series data. On the Discriminative metric, our model achieves the best or second-best results on most datasets, indicating a high degree of distributional alignment between generated and real series. This benefit can be primarily attributed to the perturbation-response mechanism, which enhances the model's sensitivity to local velocity variations. For the Predictive and Context-FID metrics, our method matches or outperforms the best-performing diffusion models, reflecting strong structural preservation and task-level generalization capabilities. This is largely due to the joint modeling of clean and perturbed trajectories, coupled with expert-guided structural adaptation in the decoder. In addition, our model consistently achieves the lowest errors on the Correlational metric, further confirming the effectiveness of the trajectory consistency block in capturing temporal coherence and dynamic dependencies. By explicitly incorporating velocity derivative alignment into the training objective, the model becomes better equipped to recover the intrinsic relational patterns.

\begin{figure}[htpb]
    \centering    
    \includegraphics[width=\linewidth]{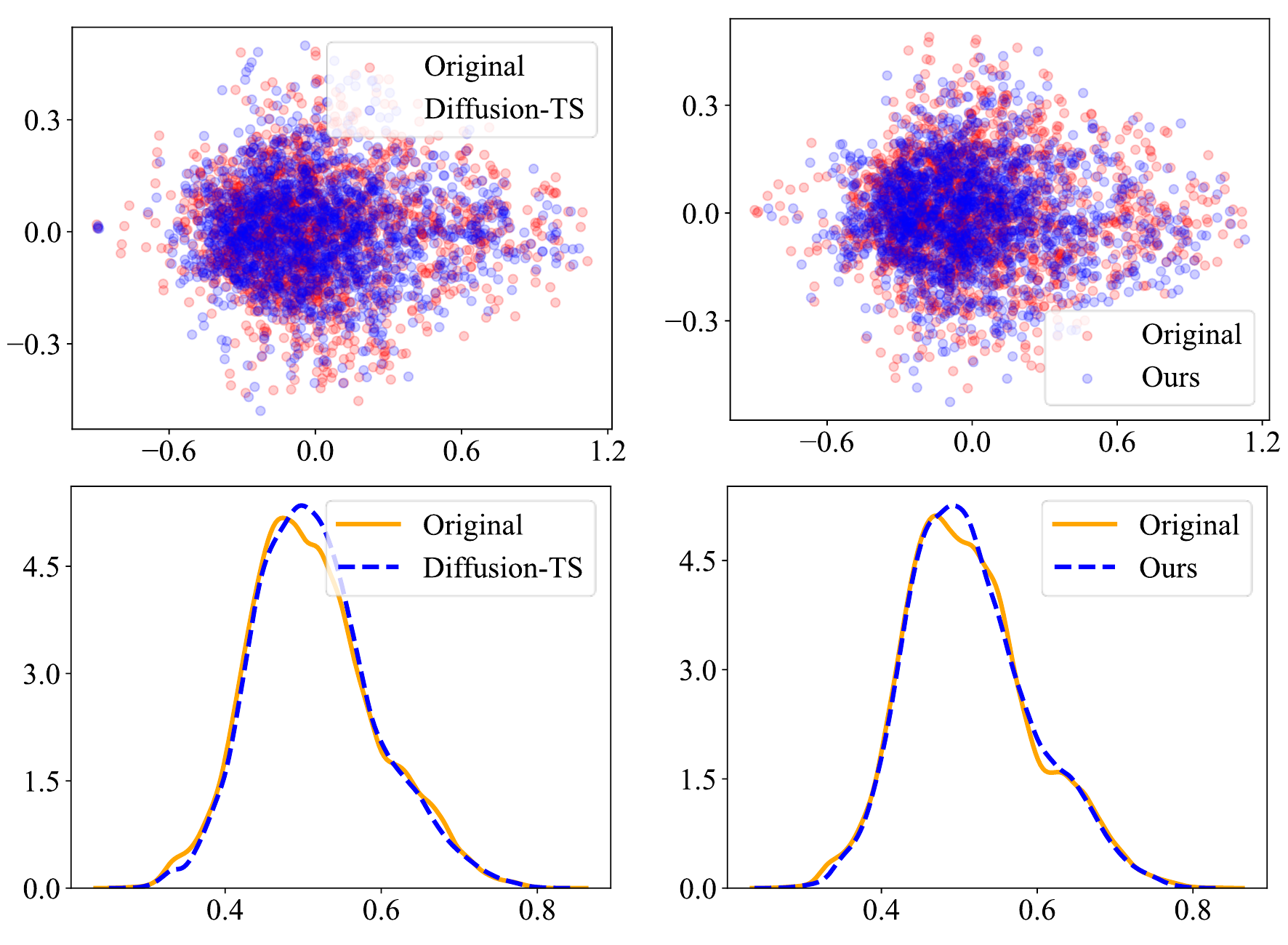}
    \vspace{-0.3in}
    \caption{Visualizations of the time series synthesized by our framework and Diffusion-TS.}
    \label{fig3}
    \vspace{-0.15in}
\end{figure}

We perform PCA-based and distributional visualizations on 10\% of the generated samples from the ETTh1 dataset. As shown in Figure~\ref{fig3}, our method achieves better alignment in both overall distribution and rare sample regions. Compared to Diffusion-TS, the generated data from our model more closely matches the real data distribution, highlighting its coherence in capturing temporal dynamics and structural dependencies. This improvement stems from the perturbation-aware modeling strategy, which enables the model to more effectively capture localized variations and maintain temporal coherence, ensuring the generated samples remain faithful to the underlying temporal and statistical data distribution.

\begin{table}[htbp]
  \centering
  \caption{Performance comparison on prediction and imputation tasks on ETTh1. Bold indicates the best performance.}
  \vspace{-0.1in}
  \label{condition}
  \small
  \resizebox{\linewidth}{!}{
  \begin{tabular}{lccccc}
    \toprule
    \textbf{Prediction (MSE ↓)} & \textbf{6} & \textbf{12} & \textbf{18} & \textbf{24} & \textbf{36} \\
    \midrule
    \cellcolor{lightgray!30}Ours         &  \cellcolor{lightgray!30}\textbf{0.001064} & \cellcolor{lightgray!30}\textbf{0.002508} & \cellcolor{lightgray!30}\textbf{0.003982} & \cellcolor{lightgray!30}\textbf{0.005495} & \cellcolor{lightgray!30}\textbf{0.011907}  \\
      Diffusion-TS   & 0.002255 & 0.004428 & 0.006212 & 0.007745 & 0.013158 \\
      CSDI           & 0.004846 & 0.009360 & 0.014213 & 0.018731 & 0.028414 \\
    \midrule
    \textbf{Imputation (MSE ↓)} & \textbf{10\%} & \textbf{25\%} & \textbf{50\%} & \textbf{75\%} & \textbf{90\%} \\
    \midrule
      \cellcolor{lightgray!30}Ours           & \cellcolor{lightgray!30}\textbf{0.000164} & \cellcolor{lightgray!30}\textbf{0.000483} & \cellcolor{lightgray!30}\textbf{0.001333} & \cellcolor{lightgray!30}\textbf{0.002885} & \cellcolor{lightgray!30}\textbf{0.002932}  \\
      Diffusion-TS   & 0.000248 & 0.000804 & 0.002164 & 0.004292 & 0.004339  \\
      CSDI           & 0.000526 & 0.002169 & 0.008920 & 0.021176 & 0.021391 \\
    \bottomrule
  \end{tabular}}
  \vspace{-0.1in}
\end{table}

\begin{table*}[htbp]
  \centering
  \caption{Ablation study for PAFM architecture. Bold indicates best performance.}
    \vspace{-0.05in}
    \begin{tabular}{c|lcccccc}
    \toprule
    Metric & Methods & Sines & Stocks & ETTh1  & MuJoCo & Energy & fMRI \\
    \midrule
    \multirow{5}[0]{*}{\begin{tabular}[c]{@{}c@{}}Discriminative\\Score\\ (Lower Better)\end{tabular}} & \cellcolor{lightgray!30}Ours  & \cellcolor{lightgray!30}\textbf{0.006±0.001} & \cellcolor{lightgray!30}\textbf{0.021±0.017} & \cellcolor{lightgray!30}\textbf{0.008±0.005} & \cellcolor{lightgray!30}\textbf{0.007±0.004} & \cellcolor{lightgray!30}\textbf{0.047±0.006} & \cellcolor{lightgray!30}0.243±0.048 \\
          & w/o FRM & 0.010±0.010 & 0.023±0.011 & 0.011± 0.006 & 0.019±0.013 & 0.056±0.006 & 0.221±0.019 \\
          & w/o TD & 0.021±0.027 & 0.281±0.092 & 0.030±0.027 & 0.216±0.093 & 0.298±0.072 & \textbf{0.029±0.026} \\
          & w/o TD\&TPB & 0.009±0.011 & 0.022±0.011 & 0.020± 0.005 & 0.047±0.098 & 0.056±0.011 & 0.140±0.072 \\
    \midrule
    \multirow{5}[0]{*}{\begin{tabular}[c]{@{}c@{}}Predictive\\Score\\(Lower Better)\end{tabular}} & \cellcolor{lightgray!30}Ours  & \cellcolor{lightgray!30}\textbf{0.093±0.000} & \cellcolor{lightgray!30}\textbf{0.036±0.000 } & \cellcolor{lightgray!30}\textbf{0.120±0.006} & \cellcolor{lightgray!30}\textbf{0.009±0.001} & \cellcolor{lightgray!30}\textbf{0.250±0.000 } & \cellcolor{lightgray!30}\textbf{0.099±0.000} \\
          & w/o FRM & \textbf{0.093±0.000} & 0.037±0.000 & 0.122±0.006 & 0.011±0.001 & 0.251±0.000 & 0.100±0.000 \\
          & w/o TD & \textbf{0.093±0.000} & 0.038±0.000 & 0.120±0.007 & 0.012±0.002 & 0.251±0.000 & 0.100±0.000 \\
          & w/o TD\&TPB & \textbf{0.093±0.000} & 0.037±0.000 & 0.122±0.003 & 0.010±0.002 & 0.251±0.000 & 0.100±0.000 \\
    \bottomrule
    \end{tabular}%
    \vspace{-0.05in}
  \label{xr}%
\end{table*}%

\subsection{Conditional Time Series Generation Results}

\begin{figure}[t!]
    \centering    
    \includegraphics[width=\linewidth]{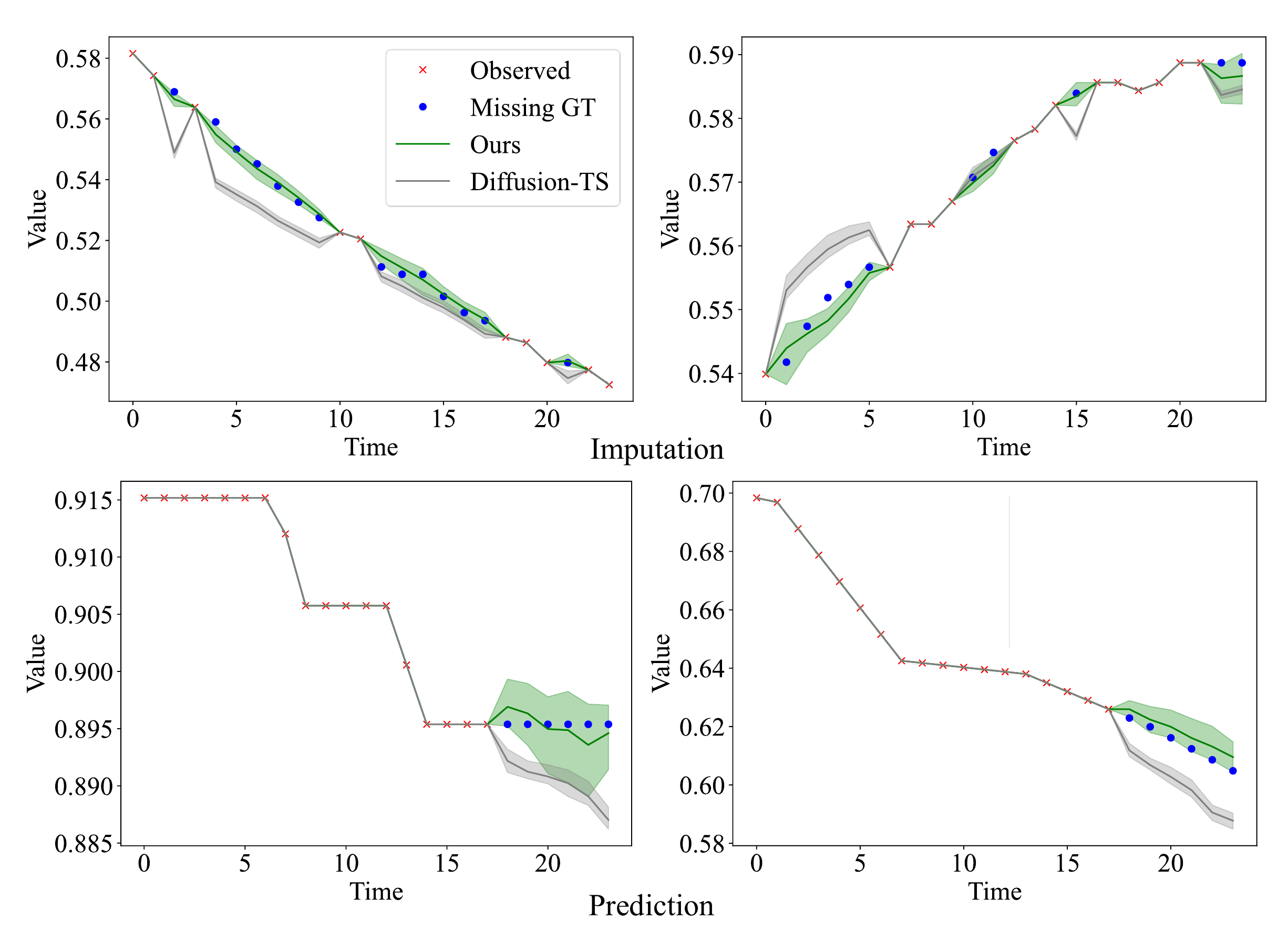}
    \vspace{-0.25in}
    \caption{Examples of time series imputation and prediction for Energy datasets. Green and gray colors correspond to our framework and Diffusion-TS, respectively.}
    \label{fig:V}
    \vspace{-0.1in}
\end{figure}

Figure~\ref{fig:V} compares our method with Diffusion-TS on the Energy dataset for both imputation and prediction tasks, with the series steps of 24 and a 50\% missing rate. Red crosses indicate observed values, blue dots represent ground truth in missing regions, and green and gray shaded areas denote the confidence intervals of our method and Diffusion-TS, respectively. In the imputation task, our method aligns more closely with the ground truth, particularly in regions with sharp local variations, demonstrating a stronger capability in modeling fine-grained temporal dynamics. In contrast, Diffusion-TS exhibits larger deviations and expanded uncertainty in several missing segments, indicating potential structural distortion. In the prediction task, our method extrapolates future trends more smoothly and directionally, with predictions that align well with the ground truth and maintain lower variance near the series boundary.

We further compare our framework with Diffusion-TS and CSDI on the ETTh1 dataset, where the total series length is fixed at 48 time steps. Given the first \( w \) observed points, the model is required to conditionally generate the remaining \( 48 - w \) steps. Performance is also evaluated under varying missing ratios to assess robustness against information sparsity. Across all settings, our framework consistently outperformed the baselines. In the imputation setting, our method maintains low error even at high missing ratios, reflecting strong resilience to extreme information sparsity. This is due to the joint modeling of perturbed and unperturbed paths, which improves uncertainty calibration and dynamic awareness. In contrast, both Diffusion-TS and CSDI degrade significantly as missing data increases. For prediction task, our framework also consistently achieves the lowest error, especially on longer horizons, where maintaining temporal consistency is more challenging. 

\vspace{-0.1in}

\subsection{Ablation Study}

To validate the functional contributions of each module, we perform ablation experiments with the time steps of 24.

\textbf{w/o FRM:} The flow routing mixture-of-experts is replaced with a standard MLP, aiming to evaluate the impact of expert-based routing on trajectory refinement.

\textbf{w/o TD:} The Trajectory Decoder is removed to examine the role of perturbation-aware mechanisms in maintaining flow.

\textbf{w/o TD \& TPB:} Both the Trajectory Decoder and Trajectory Perturbation Block are removed.

As shown in Table~\ref{xr}, we present the discriminative and predictive scores across various datasets. Replacing the flow routing mixture-of-experts with a standard MLP leads to consistent performance drops across most datasets, especially in MuJoCo and Energy. Both discriminative and predictive scores worsen, indicating that the expert-based routing mechanism enhances the adaptability to localized temporal variations and structural heterogeneity. Removing the Trajectory Decoder block results in a pronounced degradation in discriminative performance on Stocks, MuJoCo, and Energy, confirming the importance of decoding specialization for structural fidelity. Notably, the fMRI dataset exhibits a significant improvement in discriminative score when Trajectory Decoder is removed, indicating that in high-dimensional settings with complex temporal patterns, perturbation-driven supervision may play a more dominant role, potentially offsetting the impact of decoder removal. Further removing the Trajectory Perturbation Block leads to an overall deterioration in discriminative scores across datasets such as fMRI and MuJoCo. This shows that Trajectory Perturbation Block is essential for injecting targeted perturbations that reveal latent transitions and improve flow responsiveness to regime shifts. These results collectively demonstrate the complementary roles different PAFM modules. Their synergy is crucial for generating temporally coherent and structurally diverse series.

\section{Conlusion}
In this work, we proposed PAFM, a perturbation-aware flow matching framework for stable and structural temporal generation under localized perturbations and abrupt shifts. PAFM combines dual-path velocity modeling with perturbation-informed supervision, enabling localized refinement of the velocity field in response to perturbed dynamics. It further integrates expert-guided trajectory refinement to support structural specialization during generation. Trajectory perturbation experiments demonstrated the effectiveness of this strategy in revealing abrupt shifts and guiding flow correction. Furthermore, the strong performance of PAFM in unconditional generation demonstrates its ability to produce high-quality time series. Extensive evaluations on generation, prediction, and imputation tasks further show that PAFM performs consistently across varying series lengths and missing rates, indicating its effectiveness in closely approximating the realistic distribution. Ablation studies further validated the essential roles of individual components. These results highlighted the value of perturbation-aware design in enhancing the stability and temporal structural consistency of flow-based generative models. These findings highlight a promising direction for future generative modeling, where perturbation-aware mechanisms play a key role in simulating real-world disturbances and capturing localized patterns inherent in time series.




\clearpage

\bibliographystyle{ACM-Reference-Format}
\bibliography{main}

@article{flow,
  title={Flow matching for generative modeling},
  author={Lipman, Yaron and Chen, Ricky TQ and Ben-Hamu, Heli and Nickel, Maximilian and Le, Matt},
  journal={arXiv preprint arXiv:2210.02747},
  year={2022}
}

@article{adalayer1,
  title={Denoising diffusion probabilistic models},
  author={Ho, Jonathan and Jain, Ajay and Abbeel, Pieter},
  journal={Advances in neural information processing systems},
  volume={33},
  pages={6840--6851},
  year={2020}
}

@inproceedings{adalayer2,
  title={Vector quantized diffusion model for text-to-image synthesis},
  author={Gu, Shuyang and Chen, Dong and Bao, Jianmin and Wen, Fang and Zhang, Bo and Chen, Dongdong and Yuan, Lu and Guo, Baining},
  booktitle={Proceedings of the IEEE/CVF conference on computer vision and pattern recognition},
  pages={10696--10706},
  year={2022}
}

@article{diffusion-ts,
  title={Diffusion-ts: Interpretable diffusion for general time series generation},
  author={Yuan, Xinyu and Qiao, Yan},
  journal={arXiv preprint arXiv:2403.01742},
  year={2024}
}

@article{timevae,
  title={Timevae: A variational auto-encoder for multivariate time series generation},
  author={Desai, Abhyuday and Freeman, Cynthia and Wang, Zuhui and Beaver, Ian},
  journal={arXiv preprint arXiv:2111.08095},
  year={2021}
}

@article{diffwave,
  title={Diffwave: A versatile diffusion model for audio synthesis},
  author={Kong, Zhifeng and Ping, Wei and Huang, Jiaji and Zhao, Kexin and Catanzaro, Bryan},
  journal={arXiv preprint arXiv:2009.09761},
  year={2020}
}

@article{timegan,
  title={Time-series generative adversarial networks},
  author={Yoon, Jinsung and Jarrett, Daniel and Van der Schaar, Mihaela},
  journal={Advances in neural information processing systems},
  volume={32},
  year={2019}
}

@article{cotgan,
  title={Cot-gan: Generating sequential data via causal optimal transport},
  author={Xu, Tianlin and Wenliang, Li Kevin and Munn, Michael and Acciaio, Beatrice},
  journal={Advances in neural information processing systems},
  volume={33},
  pages={8798--8809},
  year={2020}
}

@article{difftime,
  title={On the constrained time-series generation problem},
  author={Coletta, Andrea and Gopalakrishnan, Sriram and Borrajo, Daniel and Vyetrenko, Svitlana},
  journal={Advances in Neural Information Processing Systems},
  volume={36},
  pages={61048--61059},
  year={2023}
}

@article{csdi,
  title={Csdi: Conditional score-based diffusion models for probabilistic time series imputation},
  author={Tashiro, Yusuke and Song, Jiaming and Song, Yang and Ermon, Stefano},
  journal={Advances in neural information processing systems},
  volume={34},
  pages={24804--24816},
  year={2021}
}

@article{metrics4,
  title={Conditional sig-wasserstein gans for time series generation},
  author={Liao, Shujian and Ni, Hao and Szpruch, Lukasz and Wiese, Magnus and Sabate-Vidales, Marc and Xiao, Baoren},
  journal={arXiv preprint arXiv:2006.05421},
  year={2020}
}

@inproceedings{GAN5,
  title={PSA-GAN: Progressive self attention GANs for synthetic time series},
  author={Jeha, Paul and Bohlke-Schneider, Michael and Mercado, Pedro and Kapoor, Shubham and Nirwan, Rajbir Singh and Flunkert, Valentin and Gasthaus, Jan and Januschowski, Tim},
  booktitle={The tenth international conference on learning representations},
  year={2022}
}

@article{GAN6,
  title={ITF-GAN: Synthetic time series dataset generation and manipulation by interpretable features},
  author={Klopries, Hendrik and Schwung, Andreas},
  journal={Knowledge-Based Systems},
  volume={283},
  pages={111131},
  year={2024},
  publisher={Elsevier}
}

@article{hypertime,
  title={Hypertime: Implicit neural representation for time series},
  author={Fons, Elizabeth and Sztrajman, Alejandro and El-Laham, Yousef and Iosifidis, Alexandros and Vyetrenko, Svitlana},
  journal={arXiv preprint arXiv:2208.05836},
  year={2022}
}

@inproceedings{DIFF2,
  title={Autoregressive denoising diffusion models for multivariate probabilistic time series forecasting},
  author={Rasul, Kashif and Seward, Calvin and Schuster, Ingmar and Vollgraf, Roland},
  booktitle={International conference on machine learning},
  pages={8857--8868},
  year={2021},
  organization={PMLR}
}

@article{DIFF4,
  title={Diffusion-based time series imputation and forecasting with structured state space models},
  author={Alcaraz, Juan Miguel Lopez and Strodthoff, Nils},
  journal={arXiv preprint arXiv:2208.09399},
  year={2022}
}

@inproceedings{DIFF5,
  title={Non-autoregressive conditional diffusion models for time series prediction},
  author={Shen, Lifeng and Kwok, James},
  booktitle={International Conference on Machine Learning},
  pages={31016--31029},
  year={2023},
  organization={PMLR}
}

@article{flow2,
  title={Flow straight and fast: Learning to generate and transfer data with rectified flow},
  author={Liu, Xingchao and Gong, Chengyue and Liu, Qiang},
  journal={arXiv preprint arXiv:2209.03003},
  year={2022}
}

@inproceedings{flow3,
  title={Scaling rectified flow transformers for high-resolution image synthesis},
  author={Esser, Patrick and Kulal, Sumith and Blattmann, Andreas and Entezari, Rahim and M{\"u}ller, Jonas and Saini, Harry and Levi, Yam and Lorenz, Dominik and Sauer, Axel and Boesel, Frederic and others},
  booktitle={Forty-first international conference on machine learning},
  year={2024}
}

@article{flow4,
  title={Lafma: A latent flow matching model for text-to-audio generation},
  author={Guan, Wenhao and Wang, Kaidi and Zhou, Wangjin and Wang, Yang and Deng, Feng and Wang, Hui and Li, Lin and Hong, Qingyang and Qin, Yong},
  journal={arXiv preprint arXiv:2406.08203},
  year={2024}
}

@article{flow5,
  title={Dirichlet flow matching with applications to dna sequence design},
  author={Stark, Hannes and Jing, Bowen and Wang, Chenyu and Corso, Gabriele and Berger, Bonnie and Barzilay, Regina and Jaakkola, Tommi},
  journal={arXiv preprint arXiv:2402.05841},
  year={2024}
}

@article{flow6,
  title={Motion flow matching for human motion synthesis and editing},
  author={Hu, Vincent Tao and Yin, Wenzhe and Ma, Pingchuan and Chen, Yunlu and Fernando, Basura and Asano, Yuki M and Gavves, Efstratios and Mettes, Pascal and Ommer, Bjorn and Snoek, Cees GM},
  journal={arXiv preprint arXiv:2312.08895},
  year={2023}
}

@article{flow8,
  title={FlowHFT: Imitation Learning via Flow Matching Policy for Optimal High-Frequency Trading under Diverse Market Conditions},
  author={Li, Yang and Chen, Zhi and Yang, Steve},
  journal={arXiv preprint arXiv:2505.05784},
  year={2025}
}

@article{flow9,
  title={Trajectory flow matching with applications to clinical time series modelling},
  author={Zhang, Xi Nicole and Pu, Yuan and Kawamura, Yuki and Loza, Andrew and Bengio, Yoshua and Shung, Dennis and Tong, Alexander},
  journal={Advances in Neural Information Processing Systems},
  volume={37},
  pages={107198--107224},
  year={2024}
}

@inproceedings{flow10,
  title={Conditional flow matching for time series modelling},
  author={Tamir, Ella and Laabid, Najwa and Heinonen, Markus and Garg, Vikas and Solin, Arno},
  booktitle={ICML 2024 Workshop on Structured Probabilistic Inference $\{$$\backslash$\&$\}$ Generative Modeling},
  year={2024}
}

@article{flowts,
  title={FlowTS: Time Series Generation via Rectified Flow},
  author={Hu, Yang and Wang, Xiao and Ding, Zezhen and Wu, Lirong and Zhang, Huatian and Li, Stan Z and Wang, Sheng and Zhang, Jiheng and Li, Ziyun and Chen, Tianlong},
  journal={arXiv preprint arXiv:2411.07506},
  year={2024}
}

@article{liuyan1,
  title={Tempo: Prompt-based generative pre-trained transformer for time series forecasting},
  author={Cao, Defu and Jia, Furong and Arik, Sercan O and Pfister, Tomas and Zheng, Yixiang and Ye, Wen and Liu, Yan},
  journal={arXiv preprint arXiv:2310.04948},
  year={2023}
}

@article{liuyan2,
  title={Interpretability and fairness evaluation of deep learning models on MIMIC-IV dataset},
  author={Meng, Chuizheng and Trinh, Loc and Xu, Nan and Enouen, James and Liu, Yan},
  journal={Scientific Reports},
  volume={12},
  number={1},
  pages={7166},
  year={2022},
  publisher={Nature Publishing Group UK London}
}

@inproceedings{y1,
  title={Robust time series analysis and applications: An industrial perspective},
  author={Wen, Qingsong and Yang, Linxiao and Zhou, Tian and Sun, Liang},
  booktitle={Proceedings of the 28th ACM SIGKDD conference on knowledge discovery and data mining},
  pages={4836--4837},
  year={2022}
}

@inproceedings{y2,
  title={Irregular traffic time series forecasting based on asynchronous spatio-temporal graph convolutional networks},
  author={Zhang, Weijia and Zhang, Le and Han, Jindong and Liu, Hao and Fu, Yanjie and Zhou, Jingbo and Mei, Yu and Xiong, Hui},
  booktitle={Proceedings of the 30th ACM SIGKDD Conference on Knowledge Discovery and Data Mining},
  pages={4302--4313},
  year={2024}
}

@inproceedings{y111,
  title={Foundation models for time series analysis: A tutorial and survey},
  author={Liang, Yuxuan and Wen, Haomin and Nie, Yuqi and Jiang, Yushan and Jin, Ming and Song, Dongjin and Pan, Shirui and Wen, Qingsong},
  booktitle={Proceedings of the 30th ACM SIGKDD conference on knowledge discovery and data mining},
  pages={6555--6565},
  year={2024}
}

@inproceedings{y4,
  title={Learning from irregularly-sampled time series: A missing data perspective},
  author={Li, Steven Cheng-Xian and Marlin, Benjamin},
  booktitle={International conference on machine learning},
  pages={5937--5946},
  year={2020},
  organization={PMLR}
}

@article{gan,
  title={Generative adversarial networks},
  author={Goodfellow, Ian and Pouget-Abadie, Jean and Mirza, Mehdi and Xu, Bing and Warde-Farley, David and Ozair, Sherjil and Courville, Aaron and Bengio, Yoshua},
  journal={Communications of the ACM},
  volume={63},
  number={11},
  pages={139--144},
  year={2020},
  publisher={ACM New York, NY, USA}
}

@misc{VAE,
  title={Auto-encoding variational bayes},
  author={Kingma, Diederik P and Welling, Max and others},
  year={2013},
  publisher={Banff, Canada}
}

@article{y6,
  title={Flow matching with gaussian process priors for probabilistic time series forecasting},
  author={Kollovieh, Marcel and Lienen, Marten and L{\"u}dke, David and Schwinn, Leo and G{\"u}nnemann, Stephan},
  journal={arXiv preprint arXiv:2410.03024},
  year={2024}
}

@article{y7,
  title={Attention is all you need},
  author={Vaswani, Ashish and Shazeer, Noam and Parmar, Niki and Uszkoreit, Jakob and Jones, Llion and Gomez, Aidan N and Kaiser, {\L}ukasz and Polosukhin, Illia},
  journal={Advances in neural information processing systems},
  volume={30},
  year={2017}
}

@article{moe,
  title={Gshard: Scaling giant models with conditional computation and automatic sharding},
  author={Lepikhin, Dmitry and Lee, HyoukJoong and Xu, Yuanzhong and Chen, Dehao and Firat, Orhan and Huang, Yanping and Krikun, Maxim and Shazeer, Noam and Chen, Zhifeng},
  journal={arXiv preprint arXiv:2006.16668},
  year={2020}
}

@article{confm,
  title={Consistency flow matching: Defining straight flows with velocity consistency},
  author={Yang, Ling and Zhang, Zixiang and Zhang, Zhilong and Liu, Xingchao and Xu, Minkai and Zhang, Wentao and Meng, Chenlin and Ermon, Stefano and Cui, Bin},
  journal={arXiv preprint arXiv:2407.02398},
  year={2024}
}

@article{adam,
  title={Adam: A method for stochastic optimization},
  author={Kingma, Diederik P and Ba, Jimmy},
  journal={arXiv preprint arXiv:1412.6980},
  year={2014}
}

@inproceedings{gfn4rec,
  title={Generative flow network for listwise recommendation},
  author={Liu, Shuchang and Cai, Qingpeng and He, Zhankui and Sun, Bowen and McAuley, Julian and Zheng, Dong and Jiang, Peng and Gai, Kun},
  booktitle={Proceedings of the 29th ACM SIGKDD Conference on Knowledge Discovery and Data Mining},
  pages={1524--1534},
  year={2023}
}

@inproceedings{TSDE,
  title={Self-supervised learning of time series representation via diffusion process and imputation-interpolation-forecasting mask},
  author={Senane, Zineb and Cao, Lele and Buchner, Valentin Leonhard and Tashiro, Yusuke and You, Lei and Herman, Pawel Andrzej and Nordahl, Mats and Tu, Ruibo and Von Ehrenheim, Vilhelm},
  booktitle={Proceedings of the 30th ACM SIGKDD Conference on Knowledge Discovery and Data Mining},
  pages={2560--2571},
  year={2024}
}

@article{deepseekmoe,
  title={Deepseekmoe: Towards ultimate expert specialization in mixture-of-experts language models},
  author={Dai, Damai and Deng, Chengqi and Zhao, Chenggang and Xu, RX and Gao, Huazuo and Chen, Deli and Li, Jiashi and Zeng, Wangding and Yu, Xingkai and Wu, Yu and others},
  journal={arXiv preprint arXiv:2401.06066},
  year={2024}
}

@inproceedings{GFN4Retention,
  title={Modeling User Retention through Generative Flow Networks},
  author={Liu, Ziru and Liu, Shuchang and Yang, Bin and Xue, Zhenghai and Cai, Qingpeng and Zhao, Xiangyu and Zhang, Zijian and Hu, Lantao and Li, Han and Jiang, Peng},
  booktitle={Proceedings of the 30th ACM SIGKDD Conference on Knowledge Discovery and Data Mining},
  pages={5497--5508},
  year={2024}
}

@inproceedings{informer,
  title={Informer: Beyond efficient transformer for long sequence time-series forecasting},
  author={Zhou, Haoyi and Zhang, Shanghang and Peng, Jieqi and Zhang, Shuai and Li, Jianxin and Xiong, Hui and Zhang, Wancai},
  booktitle={Proceedings of the AAAI conference on artificial intelligence},
  volume={35},
  number={12},
  pages={11106--11115},
  year={2021}
}

@inproceedings{y1111,
  title={8th SIGKDD International Workshop on Mining and Learning from Time Series--Deep Forecasting: Models, Interpretability, and Applications},
  author={Purushotham, Sanjay and Huan, Jun and Shen, Cong and Song, Dongjin and Wang, Yuyang and Gasthaus, Jan and Hasson, Hilaf and Park, Youngsuk and Seo, Sungyong and Nevmyvaka, Yuriy},
  booktitle={Proceedings of the 28th ACM SIGKDD Conference on Knowledge Discovery and Data Mining},
  pages={4896--4897},
  year={2022}
}

@article{cheng2025comprehensive,
  title={A comprehensive survey of time series forecasting: Concepts, challenges, and future directions},
  author={Cheng, Mingyue and Liu, Zhiding and Tao, Xiaoyu and Liu, Qi and Zhang, Jintao and Pan, Tingyue and Zhang, Shilong and He, Panjing and Zhang, Xiaohan and Wang, Daoyu and others},
  journal={Authorea Preprints},
  year={2025},
  publisher={Authorea}
}

@article{wang2025can,
  title={Can slow-thinking llms reason over time? empirical studies in time series forecasting},
  author={Wang, Jiahao and Cheng, Mingyue and Liu, Qi},
  journal={arXiv preprint arXiv:2505.24511},
  year={2025}
}

@article{luo2025time,
  title={Time Series Forecasting as Reasoning: A Slow-Thinking Approach with Reinforced LLMs},
  author={Luo, Yucong and Zhou, Yitong and Cheng, Mingyue and Wang, Jiahao and Wang, Daoyu and Pan, Tingyue and Zhang, Jintao},
  journal={arXiv preprint arXiv:2506.10630},
  year={2025}
}

@article{zhang2024conditional,
  title={Conditional Denoising Meets Polynomial Modeling: A Flexible Decoupled Framework for Time Series Forecasting},
  author={Zhang, Jintao and Cheng, Mingyue and Tao, Xiaoyu and Liu, Zhiding and Wang, Daoyu},
  journal={arXiv preprint arXiv:2410.13253},
  year={2024}
}

@article{cheng2024hmf,
  title={Hmf: A hybrid multi-factor framework for dynamic intraoperative hypotension prediction},
  author={Cheng, Mingyue and Zhang, Jintao and Liu, Zhiding and Liu, Chunli and Xie, Yanhu},
  journal={arXiv e-prints},
  pages={arXiv--2409},
  year={2024}
}

\clearpage

\appendix

\section{Experiment Details}

\subsection{baselines}

\label{baselines}

We compare our method with several representative baselines spanning diffusion-based, variational, and adversarial generative models. 
\textbf{Diffusion-TS}~\cite{diffusion-ts}~\footnote{\url{https://github.com/Y-debug-sys/Diffusion-TS}} employs a diffusion process within a Transformer backbone for conditional time series generation. 
\textbf{TimeVAE}~\cite{timevae}~\footnote{\url{https://github.com/abudesai/timeVAE}} models continuous-time latent trajectories with temporal attention to capture multi-scale dynamics. 
\textbf{Diffwave}~\cite{diffwave}~\footnote{\url{https://diffwave-demo.github.io/}} adopts a noise-conditioned diffusion framework to generate high-fidelity waveforms without adversarial training. 
\textbf{TimeGAN}~\cite{timegan}~\footnote{\url{https://github.com/jsyoon0823/TimeGAN}} integrates adversarial and supervised learning to preserve temporal dependencies and enhance latent consistency. 
\textbf{CotGAN}~\cite{cotgan}~\footnote{\url{https://github.com/tianlinxu312/cot-gan}} utilizes causal optimal transport and Sinkhorn divergence to stabilize adversarial training for series generation. 
\textbf{CSDI}~\cite{csdi}\footnote{\label{fn:csdi}\url{https://github.com/ermongroup/CSDI}} formulates time series imputation as a conditional generation task using score-based diffusion models with self-supervised learning and attention-based conditioning. 
\textbf{DiffTime}~\cite{difftime}\textsuperscript{\ref{fn:csdi}} can be regarded as an unconditional variant of CSDI, extending the score-based framework to unconditioned generation scenarios.
\textbf{FlowTS}~\cite{flowts}~\footnote{\url{https://github.com/UNITES-Lab/FlowTS}} is a rectified flow-based model that enables high-fidelity time series generation via straight-line transport in probability space.

\subsection{Metrics}

\label{metics}

\paragraph{Discriminative Score.}
To assess the realism of generated series, we adopt a discriminative evaluation metric based on binary classification~\cite{timegan}. 
A classifier is trained to distinguish real samples from synthetic ones, and the score is computed as the deviation of accuracy from a random guess baseline, specifically \( |\text{accuracy} - 0.5| \).

\paragraph{Predictive Score.}
To evaluate the prediction utility of generated data, we use a predictive score based on mean absolute error (MAE). 
A series prediction model is trained on the synthetic series and tested against ground-truth values from real data~\cite{GAN5}. 
We employ a two-layer GRU network as the predictor to ensure consistent evaluation across all baselines.

\paragraph{Contextual FID Score.}
We further evaluate the representation-level fidelity using a contextual Frechet distance. 
Specifically, we encode both real and synthetic series with a pre-trained time series encoder and compute the Frechet Inception Distance (FID) between their feature distributions~\cite{metrics4}. 
This score captures how well the synthetic data align with the real ones in representation space, and correlates with downstream task performance.

\paragraph{Correlation Structure Score.}
To compare the internal dependencies among features, we compute the pairwise covariance matrix of the time series~\cite{GAN5}. 
For features \( i \) and \( j \), the empirical covariance of real series is given by:
\begin{equation}
\widehat{\text{Cov}}_{i,j}^{(r)} = \frac{1}{T} \sum_{t=1}^{T} x_{t,i}^{(r)} x_{t,j}^{(r)} - \left( \frac{1}{T} \sum_{t=1}^{T} x_{t,i}^{(r)} \right) \left( \frac{1}{T} \sum_{t=1}^{T} x_{t,j}^{(r)} \right),
\end{equation}
and similarly for generated series \( \widehat{\text{Cov}}_{i,j}^{(g)} \). 
We then compute a normalized correlation alignment score as:
\begin{equation}
\frac{1}{d^2} \sum_{i=1}^{d} \sum_{j=1}^{d} \left| \frac{ \widehat{\text{Cov}}_{i,j}^{(r)} }{ \sqrt{ \widehat{\text{Cov}}_{i,i}^{(r)} \cdot \widehat{\text{Cov}}_{j,j}^{(r)} } } - \frac{ \widehat{\text{Cov}}_{i,j}^{(g)} }{ \sqrt{ \widehat{\text{Cov}}_{i,i}^{(g)} \cdot \widehat{\text{Cov}}_{j,j}^{(g)} } } \right|.
\end{equation}

\section{Implementation Details}

\label{implementation}

\begin{table}[htbp]
  \centering
  \caption{Hyperparameter configurations across different datasets.}
  \label{tab:hyperparam_summary}
  \small
  \resizebox{\linewidth}{!}{
    \begin{tabular}{lcccccc}
      \toprule
      \textbf{Parameter} & \textbf{Sines} & \textbf{Stocks} & \textbf{ETTh1} & \textbf{MuJoCo} & \textbf{Energy} & \textbf{fMRI} \\
      \midrule
      Attention Heads       & 4     & 4     & 4     & 4     & 4     & 4     \\
      Head Dimension          & 16    & 16    & 16    & 16    & 24    & 24    \\
      Encoder Layers        & 1     & 2     & 3     & 3     & 4     & 4     \\
      Decoder Layers        & 2     & 2     & 2     & 2     & 3     & 4     \\
      Batch Size              & 128   & 64    & 128   & 128   & 64    & 128   \\
      Sampling Steps        & 500   & 500   & 500   & 1000  & 1000  & 1000  \\
      Training Steps        & 12000 & 10000 & 18000 & 14000 & 25000 & 15000 \\
      \bottomrule
    \end{tabular}
  }
\end{table}

We report our hyperparameter configurations in table~\ref{tab:hyperparam_summary}. We perform limited hyperparameter tuning to identify a set of default configurations that achieve consistent performance across datasets. The explored search space included the following ranges: batch size~$\in \{64, 128\}$, number of attention heads~$\in \{4, 8\}$, base model dimension~$\in \{32, 64, 96, 128\}$, sampling steps~$\in \{50, 200, 500, 1000\}$. We set the perturbation magnitude \( \sigma = 0.1 \). The number of experts in the Flow Routing Mixture-of-Experts module is set to \(M=4\), with the top-2 experts selected for each trajectory segment.

\begin{table}[htbp]
  \centering
  \caption{Parameter count of different generative models.}
  \label{tab:param_comparison}
  \small
  \begin{tabular}{lccc}
    \toprule
    \textbf{Model} & \textbf{Sines} & \textbf{Stocks} & \textbf{Energy} \\
    \midrule
    TimeVAE        & 97,525    & 104,412   & 677,418   \\
    TimeGAN        & 34,026    & 48,775    & 1,043,179 \\
    Cot-GAN        & 40,133    & 52,675    & 601,539   \\
    DiffWave       & 533,592   & 599,448   & 1,337,752 \\
    Diffusion-TS   & 232,177   & 291,318   & 1,135,144 \\
    \cellcolor{lightgray!30}Ours           & \cellcolor{lightgray!30}417,540   & \cellcolor{lightgray!30}474,740   & \cellcolor{lightgray!30}1,690,000 \\
    \bottomrule
  \end{tabular}
\end{table}

We also compare our model parameter size with other baselines, as shown in Table~\ref{tab:param_comparison}.

\begin{table}[htbp]
  \centering
  \caption{Model parameter size, training time, and sampling time across datasets.}
  \label{tab:compute_resource}
  \small
  \resizebox{\linewidth}{!}{
  \begin{tabular}{lcccccc}
    \toprule
    \textbf{Compute Resource} & \textbf{Sines} & \textbf{Stocks} & \textbf{ETTh1} & \textbf{MuJoCo} & \textbf{Energy} & \textbf{fMRI} \\
    \midrule
    Model Size    & 417,540  & 474,740  & 531,940  & 534,730  & 1,690,000 & 2,090,000 \\
    Training Time (s)      & 886.93  & 747.07  & 1406.61  & 1620.00  & 3202.37   & 2104.84   \\
    Sampling Time (s)      & 19.43   & 9.73     & 42.17    & 24.47    & 157.50     & 95.48     \\
    \bottomrule
  \end{tabular}}
\end{table}

We report the computational cost of our model across six representative datasets in Table~\ref{tab:compute_resource}. Training times increase with both data scale and dimensionality, with high-dimensional datasets such as Energy and fMRI requiring substantially more computation. Despite this, the sampling process remains highly efficient, completing in under one minute across all datasets, demonstrating the practicality of our method for real-time or large-scale applications.


\section{Model Parameter Analysis}

\paragraph{Analysis of Flow Routing Mixture-of-Experts.}

\begin{table}[htbp]
  \centering
  \caption{Performance comparison under varying top-$k$ selections and number of experts $M$ across different datasets. Lower values indicate better performance.}
  \label{tab:topk_expert_comparison}
  \small
  \resizebox{\linewidth}{!}{
  \begin{tabular}{lcccccc}
    \toprule
    \textbf{Metric} & \textbf{Task} & \textbf{Setting} & \textbf{Stocks} & \textbf{ETTh1} & \textbf{Energy} & \textbf{fMRI} \\
    \midrule
    \multirow{6}{*}{Top-$k$} 
      & \multirow{3}{*}{Prediction} 
          & $k=2$ & 0.001391 & 0.006369 & 0.008327 & 0.012151  \\
      &     & $k=3$ & 0.001271 & 0.006500 & 0.008071 & 0.012249 \\
      &     & $k=4$ & 0.001559 & 0.006387 & 0.008444 & 0.012157  \\
      & \multirow{3}{*}{Imputation} 
          & $k=2$ & 0.00002661 & 0.00016134 & 0.00024411 & 0.00172999 \\
      &     & $k=3$ & 0.00002537 & 0.00015749 & 0.00022306 & 0.00172222 \\
      &     & $k=4$ & 0.00002769 & 0.00016174 & 0.00024130 & 0.00173221  \\
    \midrule
    \multirow{6}{*}{$E$} 
      & \multirow{3}{*}{Prediction} 
          & $M=4$ & 0.001391 & 0.006369 & 0.008327 & 0.012151  \\
      &     & $M=5$ & 0.000721 & 0.006491 & 0.008363 & 0.012276 \\
      &     & $M=6$ & 0.001784 & 0.006674 & 0.008452 & 0.012280 \\
      & \multirow{3}{*}{Imputation} 
          & $M=4$ & 0.00002661 & 0.00016134 & 0.00024411 & 0.00172999  \\
      &     & $M=5$ & 0.00002352 & 0.00016045 & 0.00023196 & 0.00171081 \\
      &     & $M=6$ & 0.00002565 & 0.00017034 & 0.00025413 & 0.00172331 \\
    \bottomrule
  \end{tabular}}
\end{table}


To evaluate the Flow Routing Mixture-of-Experts (FR-MoE), we vary either the number of selected experts ($k$) or the total number of experts ($M$), keeping the other fixed ($M{=}4$ for top-$k$ evaluation; $k{=}2$ for expert size evaluation). As shown in Table~\ref{tab:topk_expert_comparison}, increasing $k$ from 2 to 3 improves prediction performance across datasets, while $k{=}4$ shows marginal or no further gain. In contrast, smaller $k$ values yield better imputation accuracy, likely due to reduced redundancy. Increasing $M$ from 4 to 5 improves performance in both tasks, but $M{=}6$ shows saturation or degradation. These results indicate that balancing sparsity and capacity is critical for expert specialization, especially under partial observations.

\paragraph{Effect of Different Steps.}

\begin{figure}[htpb]
    \centering    
    \includegraphics[width=\linewidth]{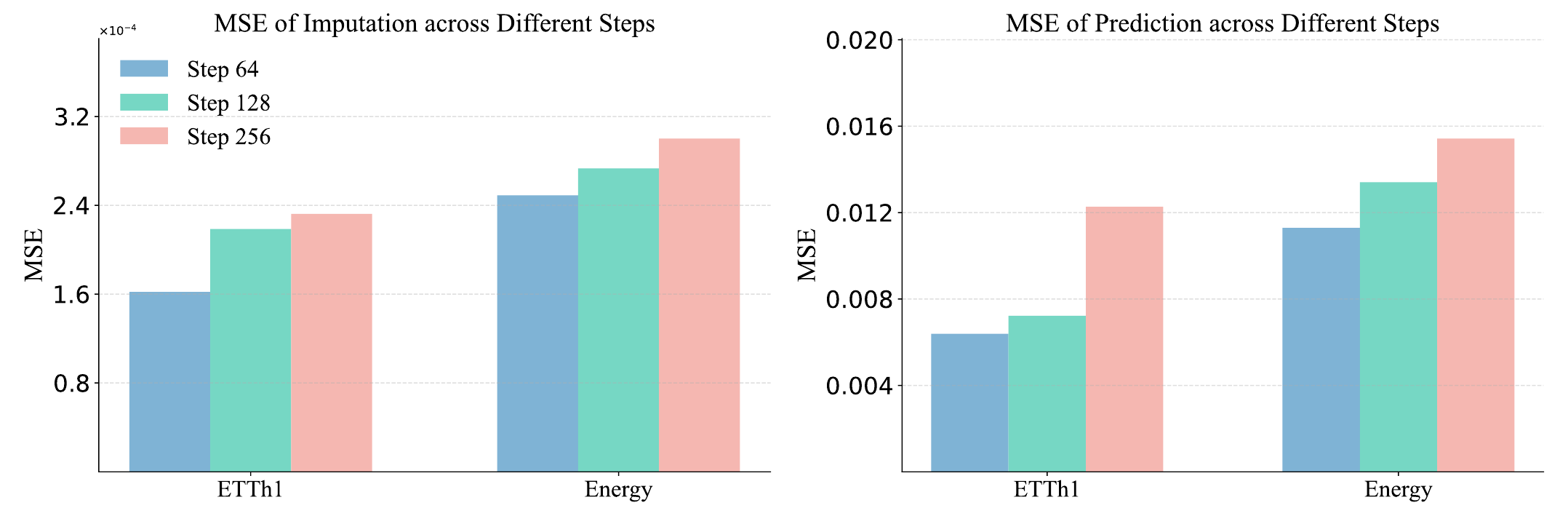}
    \caption{Imputation and prediction results on the ETTh1 and Energy datasets across different generation steps.}
    \label{fig:step_analysis}
\end{figure}

We investigate the effect of varying integration step sizes on imputation and prediction tasks, as illustrated in Fig.~\ref{fig:step_analysis}. In the imputation task, the model preserves local structure across different granularities, indicating effective adaptation to integration coarseness. In the prediction task, it maintains temporal dynamics and trend fidelity under a longer prediction horizon. These results demonstrate the model's ability to represent trajectories under varying temporal abstraction while preserving structural consistency and directional alignment.

\section{Algorithm Details}
The training and inference procedures are summarized in Algorithm~\ref{alg:abstract-train} and Algorithm~\ref{alg:abstract-infer}, respectively.

\paragraph{Training.}
The model minimizes the loss with perturbation-guided refinement. At each timepoint \( t \in [0, 1] \), a perturbed midpoint \( x_t' \) is generated through perturbation. The velocity field is evaluated on both \( x_t \) and \( x_t' \), with correction and finite-difference alignment to enhance stability and temporal smoothness.

\begin{algorithm}[H]
\caption{Training of Perturbation-Aware Flow Matching}
\label{alg:abstract-train}
\begin{algorithmic}[1]
\REQUIRE Dataset $\mathcal{D} = \{(x_0, x_1)\}$, velocity field $v_\theta$, perturbation scale $\sigma$, consistency weight $\alpha_\mathcal{L}$

\FOR{each minibatch $(x_0, x_1) \sim \mathcal{D}$}
    \STATE Sample $t \sim \mathcal{U}(0, 1)$
    \STATE $x_t \gets (1 - t)x_0 + t x_1$
    \STATE $x_t' \gets x_t + \sigma \cdot \epsilon$, where $\epsilon \sim \mathcal{N}(0, I)$

    \STATE $v \gets v_\theta(x_t, t)$,\quad $v' \gets v_\theta(x_t', t)$
    \STATE $\delta \gets v' - v$ \hfill // perturbation response
    \STATE $\hat{v} \gets v + \alpha \cdot \delta \cdot w(v)$ \hfill // refined velocity

    \STATE $\mathcal{L}_{\text{fm}} \gets \| \hat{v} - (x_1 - x_0) \|^2$
    
    \STATE Update $\theta$ by minimizing $\mathcal{L}_{\text{fm}}$
\ENDFOR
\end{algorithmic}
\end{algorithm}

\paragraph{Inference.}
Inference integrates the learned velocity field from an initial noise \( x_0 \sim \mathcal{N}(0, I) \) using an Euler scheme. Perturbation refinement improves geometric sensitivity. The process is deterministic and enables smooth, structure-aware generation. The integration step size $\Delta t$ is generally defined as a uniform temporal interval between consecutive time points, such that $\Delta t = t_{i+1} - t_i$.

\begin{algorithm}[H]
\caption{Trajectory Generation via Velocity Field Integration}
\label{alg:abstract-infer}
\begin{algorithmic}[1]
\REQUIRE Initial state $x_0$, velocity field $v_\theta$, time steps $\{t_0, t_1, \dots, t_T\}$

\STATE $x_{t_0} \gets x_0$
\FOR{$i = 0$ to $T - 1$}
    \STATE $v \gets v_\theta(x_{t_i}, t_i)$
    \STATE  $\epsilon \sim \mathcal{N}(0, I)$,\quad $x_{t_i}' \gets x_{t_i} + \sigma\cdot \epsilon$
    \STATE  $v' \gets v_\theta(x_{t_i}', t_i)$
    \STATE  $v \gets v + \alpha \cdot (v' - v) \cdot w(v)$
    \STATE $x_{t_{i+1}} \gets x_{t_i} + \Delta t \cdot v$
\ENDFOR
\STATE \textbf{return} $x_{t_T}$
\end{algorithmic}
\end{algorithm}

\section{Additional Visualization Results}

\begin{figure*}
    \centering    
    \includegraphics[width=\linewidth]{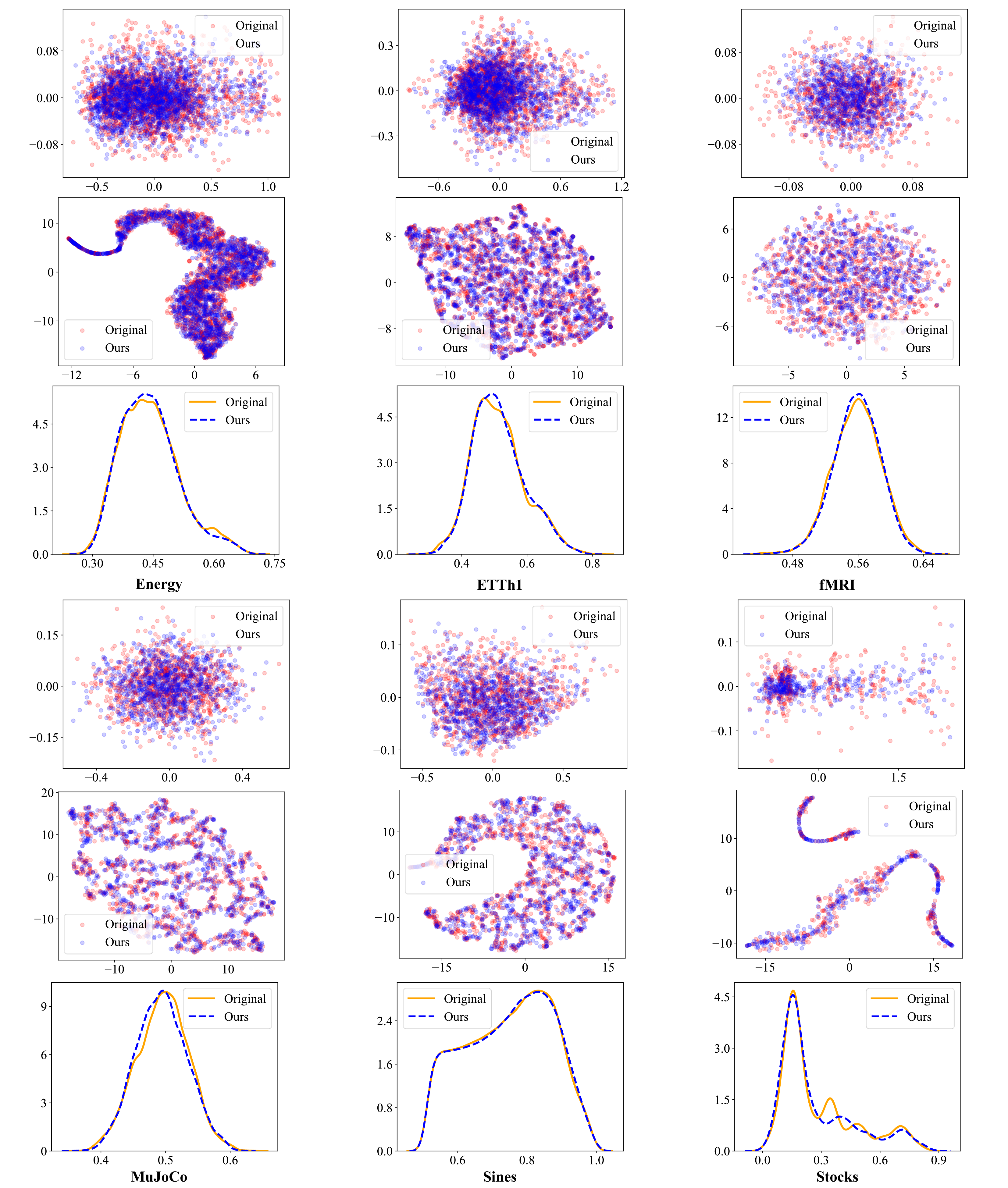}
    \caption{PCA, t-SNE, and value distribution plots of original and generated samples.}
    \label{fig:vis_pca_tsne_dis}
\end{figure*}

\begin{figure*}
    \centering    
    \includegraphics[width=\linewidth]{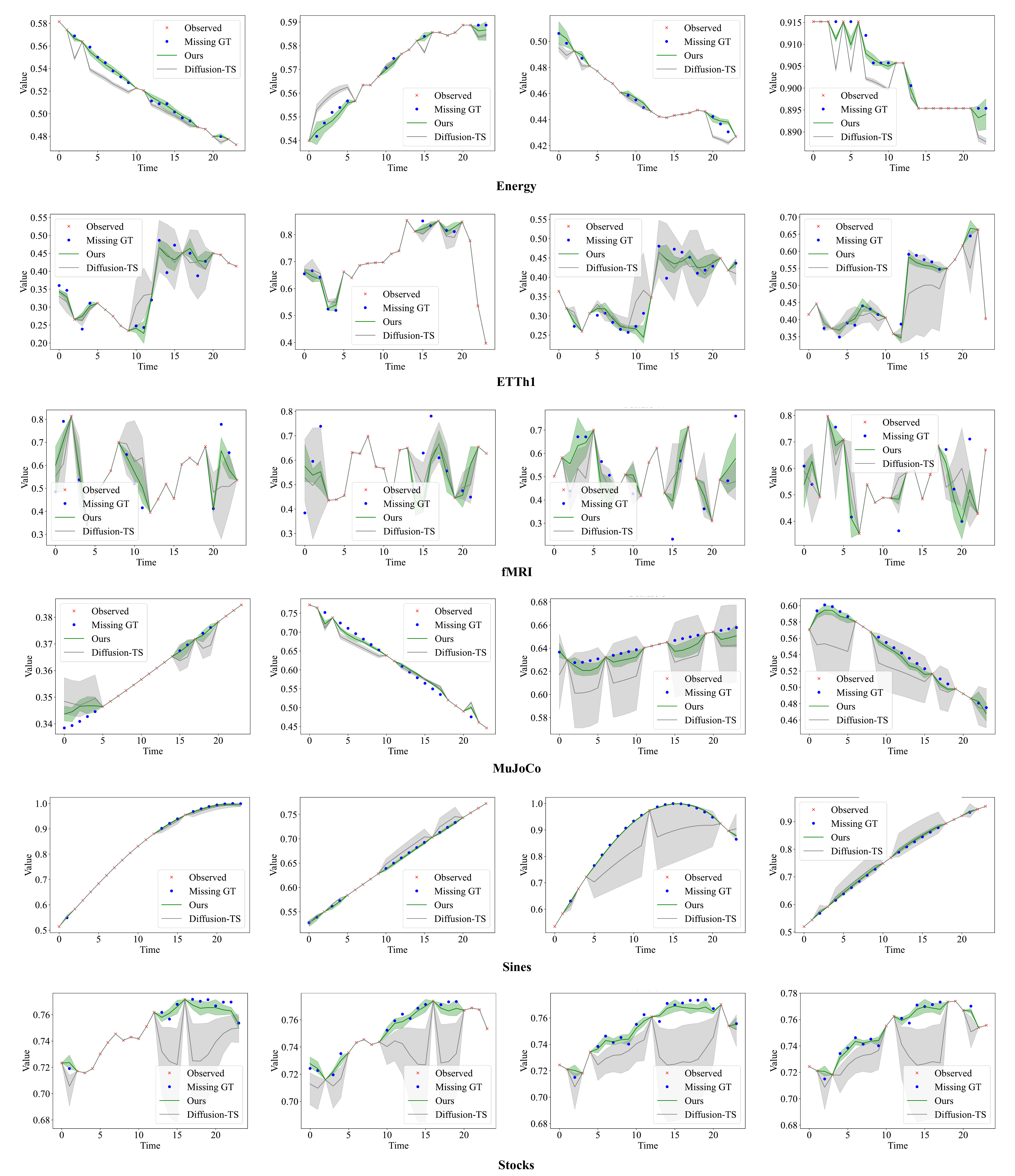}
    \caption{Visualization of imputation results across different datasets.}
    \label{fig:vis_imp}
\end{figure*}

\begin{figure*}
    \centering    
    \includegraphics[width=\linewidth]{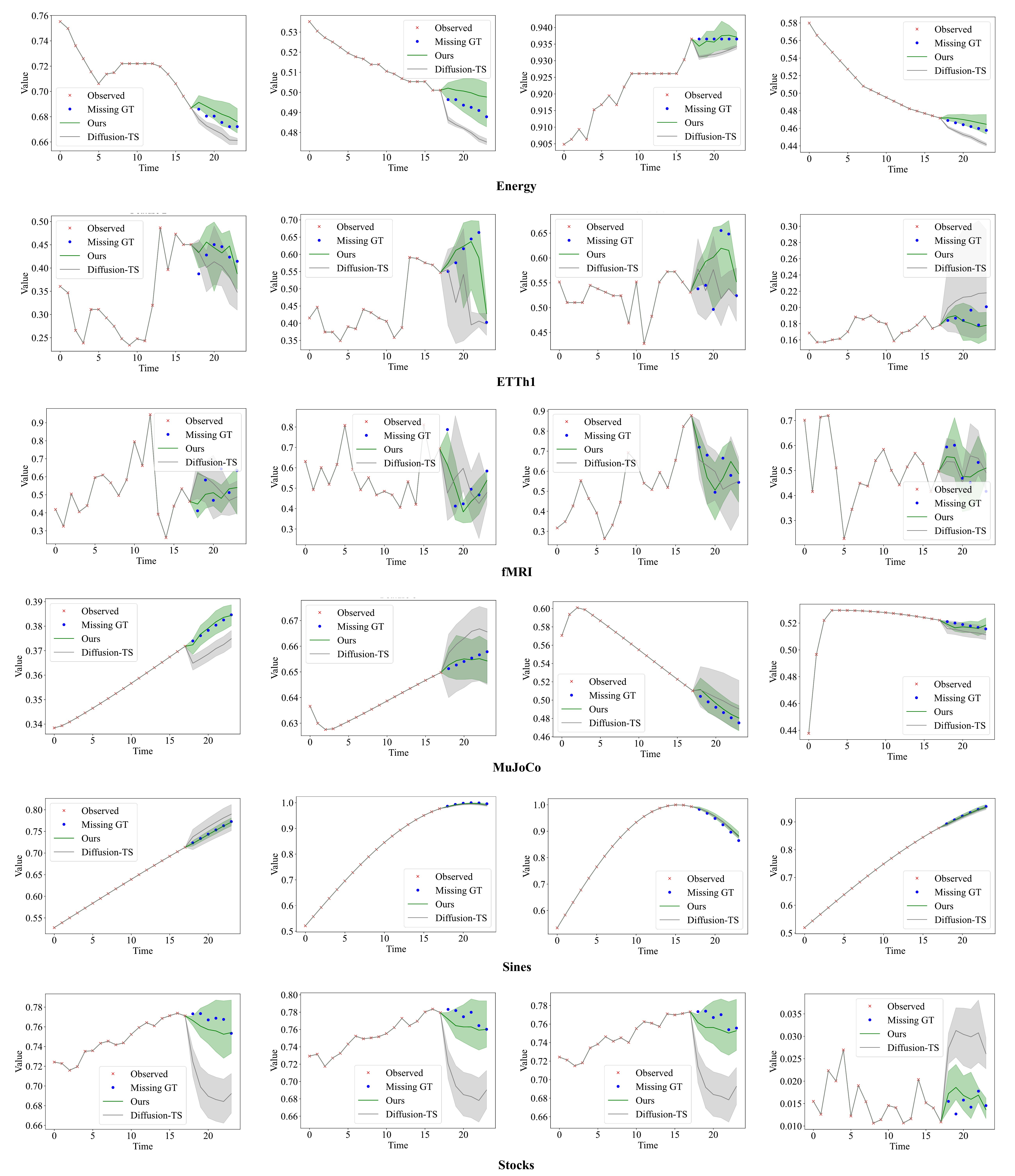}
    \caption{Visualization of prediction results across different datasets.}
    \label{fig:vis_pred}
\end{figure*}

We randomly select 10\% of the generated samples for analysis. As shown in Fig.~\ref{fig:vis_pca_tsne_dis}, both PCA and t-SNE projections indicate that the generated series preserve the geometric structure of the original data. Value distribution plots further show strong statistical alignment, suggesting that global dynamics and local variations are well captured. As shown in Fig.~\ref{fig:vis_imp}, our model outperforms the diffusion baseline under a 50\% missing rate, particularly in regions with sharp transitions. Predictions align more closely with the ground truth and exhibit tighter uncertainty bands, reflecting improved confidence and reconstruction quality. Fig.~\ref{fig:vis_pred} shows that our model extrapolates both smooth and volatile patterns more accurately than the diffusion baseline. It yields narrower, better-calibrated uncertainty bands, especially in perturbed regions, demonstrating strong modeling of complex dynamics under limited observations.




\end{document}